\documentclass[10pt]{article}
\usepackage{palatino}
\usepackage{color, palatino}
\usepackage[top=3cm, bottom=3cm, left=2cm, right=2cm]{geometry}

\usepackage{amsmath}
\usepackage[ansinew]{inputenc}
\usepackage{bm}
\usepackage{float}
\usepackage{url}
\usepackage{algorithm}
\usepackage{algorithmic}
\usepackage{graphics,graphicx}
\usepackage{subfigure}
\usepackage{color}
\usepackage{amsmath}
\usepackage[ansinew]{inputenc}
\usepackage{bm}
\usepackage{float}
\usepackage{url}
\usepackage{algorithm}
\usepackage{algorithmic}
\usepackage{graphics}
\usepackage{subfigure}
\usepackage{hyperref}
\usepackage[all]{hypcap}   

\DeclareMathOperator{\tr}{trace}           
\DeclareMathOperator{\cov}{cov}         
\DeclareMathOperator{\ex}{E}            
\DeclareMathOperator{\diag}{diag}       

\newcommand{\rbfKernel}{EQ}
\newcommand{\rbfKernelLong}{exponentiated quadratic}
\newcommand{\boldx}{\mathbf{x}} 
\newcommand{\boldc}{\mathbf{c}} 
\newcommand{\boldcv}{\overline{\mathbf{c}}} 
\newcommand{\boldyv}{\overline{\mathbf{y}}} 
\newcommand{\boldw}{\mathbf{w}} 
\newcommand{\boldz}{\mathbf{z}} 
\newcommand{\boldh}{\mathbf{h}} 
\newcommand{\boldk}{\mathbf{k}} 
\newcommand{\Kms}{k(\boldX,\boldX)} 
\newcommand{\Kmsd}[2]{(\vk(\boldX_{{#1}},\boldX_{{#2}}))_{{#1},{#2}}}
\newcommand{\Kmv}{\vk(\boldX,\boldX)} 
\newcommand{\boldS}{\mathbf{S}} 
\newcommand{\boldK}{\mathbf{K}} 
\newcommand{\boldL}{\mathbf{L}} 
\newcommand{\boldD}{\mathbf{D}} 
\newcommand{\boldC}{\mathbf{C}} 
\newcommand{\boldf}{\mathbf{f}} 
\newcommand{\bolde}{\mathbf{e}} 
\newcommand{\fX}{f(\boldX)} 
\newcommand{\FX}{\boldf(\boldX)} 
\newcommand{\boldB}{\mathbf{B}} 
\newcommand{\boldM}{\mathbf{M}} 
\newcommand{\boldE}{\mathbf{E}} 
\newcommand{\boldG}{\mathbf{G}} 
\newcommand{\boldZ}{\mathbf{Z}} 
\newcommand{\boldu}{\mathbf{u}} 
\newcommand{\boldy}{\mathbf{y}} 
\newcommand{\bolda}{\mathbf{a}} 
\newcommand{\eye}{\mathbf{I}}   
\newcommand{\boldI}{\mathbf{I}}   
\newcommand{\boldAtilde}{\mathbf{\widetilde{A}}} 
\newcommand{\boldX}{\mathbf{X}} 
\newcommand{\dif}{\textrm{d}}   
\newcommand{\inputSpace}{\mathcal{X}} 
\newcommand{\rone}{{\mathbf R}}
\newcommand{\vk}{{\mathbf K}}
\newcommand{\SX}{\mathbf{X}}
\newcommand{\SY}{\mathbf{Y}}
\newcommand{\gauss}{\mathcal{N}} 

\newcommand{\new}{\newcommand}
\new{\bg}{\begin}
\new{\lp}{\left(}
\new{\rp}{\right)}

\new{\iii}{\begin{enumerate}}
\new{\fff}{\end{enumerate}}
\new{\iiii}{\begin{itemize}}
\new{\ffff}{\end{itemize}}
\new{\mfi}{\begin{eqnarray*}}
\new{\mff}{\end{eqnarray*}}
\new{\mfni}{\begin{eqnarray}}
\new{\mfnf}{\end{eqnarray}}
\new{\beeq}[2]{\begin{equation}\label{#1}{#2}\end{equation}}
\new{\eqn}[1]{(\ref{#1})}

\new{\room}{\ \ \ \ }
\new{\card}{\#}
\newcommand{\1}{\mathbf{1}}
\newcommand{\Id}{\mathbf{I}}

\new{\nor}[1]{\|{#1}\|}
\new{\normi}[1]{\left\|{#1}\right\|_{\infty}}
\new{\scal}[2]{\left\langle{#1},{#2}\right\rangle}
\new{\scalh}[2]{\left\langle{#1},{#2}\right\rangle_\hh}
\new{\set}[1]{\{{#1}\}}
\new{\com}{{\mathbb C}}
\new{\nat}{{\mathbb N}}

\new{\fz}{f_n}
\new{\fzlo}{\overline{f}_\vz^\la}
\new{\rd}{\rone^\dm}
\new{\ed}{{\mathbb E}}

\new{\argmin}{\operatornamewithlimits{argmin}}
\new{\argmax}{\operatornamewithlimits{argmax}}
\new{\Prob}[1]{\mathrm{P}\left[\, #1 \right]}
\new{\E}{{\mathbb E}}
\new{\eps}{\varepsilon}

\new{\marg}{\rho_X}
\new{\prob}{\rho}
\new{\dm}{d}
\new{\n}{n}
\new{\vb}{{\mathbf b}}
\new{\vf}{{\mathbf f}}
\new{\vl}{\bar{\ell}}
\new{\vx}{{\mathbf x}}
\new{\vy}{{\mathbf y}}
\new{\vz}{{\mathbf z}}
\new{\vc}{{\mathbf c}}
\new{\vu}{{\mathbf u}}
\new{\vv}{{\mathbf v}}
\new{\kk}{K}
\new{\s}{K^s}
\new{\kkx}{K_x}
\new{\w}{W}
\new{\wx}{\w_x}
\new{\WW}{{\mathbf W}}
\new{\DD}{{\mathbf D}}
\new{\LL}{{\mathbf L}}
\new{\LS}{{\mathbf L}_{s}}
\new{\LR}{{\mathbf L}_{r}}
\new{\II}{{\mathbf I}}
\new{\KK}{{\mathbf K}}
\new{\ka}{\kappa}
\new{\ls}{\ell}
\new{\err}{{\mathcal E}}
\new{\emp}{{\mathcal E}_\ts}
\new{\hh}{{\mathcal H}}
\new{\oo}{{\mathcal G}}
\new{\ff}{{\mathcal F}}
\new{\ldue}{L^2(X,\rho)}
\new{\la}{\lambda}
\new{\dd}{\delta}
\new{\vG}{{\mathbf \Gamma}}
\new{\vC}{{\mathbf C}}
\new{\vS}{{\mathbf S}}
\new{\vU}{{\mathbf U}}
\new{\vY}{{\mathbf Y}}
\new{\vX}{{\mathbf X}}
\new{\vaa}{{\boldsymbol {\alpha}}}
\new{\bh}{{\mathcal B}(\hh)}
\new{\hs}{ HS(\hh)}

\new{\lduew}{L^2(X,\rho_\w)}
\new{\Tt}{L_{K,\hh}}
\new{\Tn}{L_{K,n}}
\new{\U}{U_\w}
\new{\LK}{L_{K}}

\new{\Wt}{L_{\w,\hh}}
\new{\Wn}{L_{\w,n}}

\new{\Ah}{A_{\hh}}
\new{\An}{A_{n}}

\new{\Ls}{L_{s}}
\new{\Lr}{L_{r}}

\new{\Lshn}{L_{s,\hh,n}}
\new{\Lrhn}{L_{r,\hh,n}}
\new{\LLsh}{L_{s,\hh}}
\new{\LLrh}{L_{r,\hh}}
\new{\I}{I_{\kk,\w}}
\new{\Ik}{I_\kk}
\new{\IIk}{I^*_\kk}
\new{\Iw}{I_\w}
\new{\IIw}{I^*_\w}

\new{\m}{m}
\new{\mn}{m_{n}}

\new{\Ax}{{A_{\mathbf x}}}
\new{\A}{A}
\new{\AR}{\A^\la}
\new{\ARx}{\A^\la}
\new{\Tx}{L_{K,n}}
\new{\Sx}{{S_n}}
\new{\T}{T}
\new{\Kx}{{K_{\mathbf x}}}
\new{\K}{K}
\new{\g}{{\mathcal G}}
\new{\sn}{\hat{\sigma}}
\new{\vn}{\hat{v}}
\new{\un}{\hat{u}}

\new{\so}{{s^\prime}}

\new{\Pj}{P_j}
\new{\Qj}{Q_j}
\new{\Pnj}{\hat{P}_j}
\new{\Qnj}{\hat{Q}_j}

\new{\cl}{c}
\new{\br}{b}
\new{\R}{R}
\new{\vre}{\vf_\rho}
\new{\re}{f_\rho}
\new{\X}{\mathcal X}
\new{\Y}{\mathcal Y}
\new{\W}{\mathcal W}

\begin{document}

%


\title{Kernels for Vector-Valued Functions: a Review}
\author{Mauricio A. \'Alvarez$^{+}$,  Lorenzo Rosasco$^{\sharp,\dagger}$, Neil D. Lawrence$^{\star,\diamond}$,   \\
\small \em $\ddagger$ - School of Computer Science, University of Manchester Manchester, UK, M13 9PL.\\
\small \em $+$ Department of Electrical Engineering, Universidad Tecnol\'ogica de Pereira, Colombia, 660003\\
\small \em $\sharp$ - CBCL, McGovern Institute, Massachusetts Institute of Technology, Cambridge, MA, USA\\
\small \em $\dagger$ -IIT@MIT Lab,  Istituto Italiano di Tecnologia,   Genova, Italy\\
\small \em $\star$ - Department of Computer Science, University of Sheffield, UK \\
\small \em $\diamond$ The Sheffield Institute for Translational Neuroscience, Sheffield, UK.\\
\small \tt malvarez@utp.edu.co, lrosasco@mit.edu, n.lawrence@sheffield.ac.uk}

\date \today

\maketitle

\abstract{Kernel methods are among the most popular techniques in
  machine learning. From a regularization perspective they play a
  central role in regularization theory as they provide a natural
  choice for the hypotheses space and the regularization functional
  through the notion of reproducing kernel Hilbert spaces.  From a
  probabilistic perspective they are the key in the context of
  Gaussian processes, where the kernel function is known as the
  covariance function.  Traditionally, kernel methods have been used
  in supervised learning problem with scalar outputs and indeed there
  has been a considerable amount of work devoted to designing and
  learning kernels.  More recently there has been an increasing
  interest in methods that deal with multiple outputs, motivated
  partly by frameworks like multitask learning. In this paper, we
  review different methods to design or learn valid kernel functions
  for multiple outputs, paying particular attention to the connection
  between probabilistic and functional methods. }

\newpage

\tableofcontents

\newpage

\section{Introduction}

Many modern applications of machine learning require solving several
decision making or prediction problems and exploiting dependencies
between the problems is often the key to obtain better results and
coping with a lack of data (to solve a problem we can {\em borrow
  strength} from a distinct but related problem).

In \emph{sensor networks}, for example, missing signals from certain
sensors may be predicted by exploiting their correlation with observed
signals acquired from other sensors \cite{Rogers:towards08}. In
\emph{geostatistics}, predicting the concentration of heavy pollutant
metals, which are expensive to measure, can be done using inexpensive
and oversampled variables as a proxy \cite{Goovaerts:book97}. In
\emph{computer graphics}, a common theme is the animation and
simulation of physically plausible humanoid motion. Given a set of
poses that delineate a particular movement (for example, walking), we
are faced with the task of completing a sequence by filling in the
missing frames with natural-looking poses.  Human movement exhibits a
high-degree of correlation. Consider, for example, the way we
walk. When moving the right leg forward, we unconsciously prepare the
left leg, which is currently touching the ground, to start moving as
soon as the right leg reaches the floor. At the same time, our hands
move synchronously with our legs.  We can exploit these implicit
correlations for predicting new poses and for generating new
natural-looking walking sequences \cite{Wang:GPDM:2008}.  In
\emph{text categorization}, one document can be assigned to multiple
topics or have multiple labels \cite{Kazawa:multiTopicTextCat:2005}.
In all the examples above, the simplest approach ignores the potential
correlation among the different output components of the problem and
employ models that make predictions individually for each output.
However, these examples suggest a different approach through a joint
prediction exploiting the interaction between the different components
to improve on individual predictions.  Within the machine learning
community this type of modeling is often broadly referred to to as
\emph{multitask learning}.  Again the key idea is that information
shared between different tasks can lead to improved performance in
comparison to learning the same tasks individually. These ideas are
related to {\em transfer learning} \cite{Thrun:learningnthing:1996,
  Caruana:MTL:1997, Bonilla:multi07, PQ10}, a term which refers to
systems that learn by transferring knowledge between different
domains, for example: ``what can we learn about running through seeing
walking?''

More formally, the classical supervised learning problem requires
estimating the output for any given input $\boldx_*$; an estimator
$f_*(\boldx_*)$ is built on the basis of a training set consisting of
$N$ input-output pairs $S=(\SX, \SY)=(\boldx_1,y_1),\dots,
(\boldx_N,y_N)$.  The input space $\mathcal{X}$ is usually a space of
vectors, while the output space is a space of {\em scalars}.  In
multiple output learning (MOL) the output space is a space of {\em
  vectors}; the estimator is now a {\em vector valued function}
$\boldf$.  Indeed, this situation can also be described as the problem
of solving $D$ distinct classical supervised problems, where each
problem is described by one of the components $f_1, \dots, f_D$ of
$\boldf$.  As mentioned before, the key idea is to work under the
assumption that the problems are in some way related.  The idea is
then to exploit the relation among the problems to improve upon
solving each problem separately.

The goal of this survey is twofold. First, we aim at discussing recent
results in multi-output/multi-task learning based on kernel methods
and Gaussian processes providing an account of the state of the art in
the field.  Second, we analyze systematically the connections between
Bayesian and regularization (frequentist) approaches. Indeed, related
techniques have been proposed from different perspectives and drawing
clearer connections can boost advances in the field, while fostering
collaborations between different communities.

The plan of the paper follows.  In chapter \ref{Sec:ScalLearn} we give
a brief review of the main ideas underlying kernel methods for scalar
learning, introducing the concepts of regularization in reproducing
kernel Hilbert spaces and Gaussian processes.  In chapter
\ref{chap:VecLearn} we describe how similar concepts extend to the
context of vector valued functions and discuss different settings that
can be considered.  In chapters \ref{SepKer} and \ref{NonSepKer} we
discuss approaches to constructing multiple output kernels, drawing
connections between the Bayesian and regularization frameworks. The
parameter estimation problem and the computational complexity problem
are both described in chapter \ref{parEstimation}. In chapter
\ref{applications} we discuss some potential applications that can be
seen as multi-output learning.  Finally we conclude in chapter
\ref{Conc} with some remarks and discussion.

\section{Learning Scalar Outputs with Kernel Methods}\label{Sec:ScalLearn}

To make the paper self contained, we will start our study reviewing
the classical problem of learning a scalar valued function, see for
example \cite{vapnik98,Hastie:Elements:2009, Bishop:PRLM06,
  Rasmussen:book06}.  This will also serve as an opportunity to review
connections between Bayesian and regularization methods.

As we mentioned above, in the classical setting of supervised
learning, we have to build an estimator (e.g. a classification rule or
a regression function) on the basis of a training set $S=(\SX,
\SY)=(\boldx_1,y_1),\dots, (\boldx_N,y_N)$.  Given a symmetric and
positive bivariate function $k(\cdot, \cdot)$, namely a {\em kernel},
one of the most popular estimators in machine learning is defined as
\begin{equation}\label{TikhoEsti}
f_*(\boldx_*)={\mathbf k}_{\boldx_*}^{\top}(\Kms+\la N\eye )^{-1} \SY,
\end{equation}
where $\Kms$ has entries $k(\boldx_i,\boldx_j)$, $\SY=[y_1, \dots,
y_N]^{\top}$ and ${\mathbf k}_{\boldx_*}=
[k(\boldx_1,\boldx_*),\dots,k(\boldx_N,\boldx_*)]^{\top}$, where
$\boldx_*$ is a new input point.  Interestingly, such an estimator can
be derived from two different, though, related perspectives.

\subsection{A Regularization  Perspective}

We will first describe a regularization (frequentist) perspective (see
\cite{girpog89,Wahba:splinesBook:1990,vapnik98,Scholkopf:kernelsBook:2002}).
The key point in this setting is that the function of interest is
assumed to belong to a reproducing kernel Hilbert space (RKHS),
$$f_*\in \hh_k.$$
Then the estimator is derived as the minimizer of a regularized
functional
\begin{align}\label{tikho}
\frac{1}{N}\sum_{i=1}^N (f(\boldx_i)-y_i)^2+\la \nor{f}^2_k.
\end{align}
The first term in the functional is the so called empirical risk and
it is the sum of the squared errors.
It is a measure of the price we pay when predicting $f(\boldx)$ in
place of $y$.  The second term in the functional is the (squared) norm
in a RKHS.  This latter concept plays a key role, so we review a few
essential concepts (see \cite{schwartz, aron,
  Wahba:splinesBook:1990,cucsma02}).  A RKHS $\hh_k$ is a Hilbert
space of functions and can be defined by a reproducing
kernel\footnote{In the following we will simply write kernel rather
  than reproducing kernel.} $k:\mathcal{X}\times \mathcal{X}\to
\rone$, which is a symmetric, positive definite function.  The latter
assumption amounts to requiring the matrix with entries
$k(\boldx_i,\boldx_j)$ to be positive for any (finite) sequence
$(\boldx_i)$.  Given a kernel $k$, the RKHS $\hh_k$ is the Hilbert
space such that the function $k(\boldx,\cdot)$ belongs to belongs to
$\hh_k$ for all $\boldx\in \mathcal{X}$ and
$$
f(\boldx)=\scal{f}{k(\boldx,\cdot)}_k,\quad \forall~f\in \hh_k,
$$
where $\scal{\cdot}{\cdot}_k$ is the inner product in $\hh_k$.

The latter property, known as the reproducing property, gives the name
to the space.  Two further properties make RKHS appealing:
\begin{itemize}
\item functions in a RKHS are in the closure of the linear
  combinations of the kernel at given points, $f(\boldx)= \sum_i
  k(\boldx_i,\boldx)c_i$.  This allows us to describe, in a unified
  framework, linear models as well as a variety of generalized linear
  models;
\item the norm in a RKHS can be written as $\sum_{i,j}k(\boldx_i,
  \boldx_j)c_ic_j$ and is a natural measure of how {\em complex} is a
  function. Specific examples are given by the shrinkage point of view
  taken in ridge regression with linear models
  \cite{Hastie:Elements:2009} or the regularity expressed in terms of
  magnitude of derivatives, as is done in spline models
  \cite{Wahba:splinesBook:1990}.
\end{itemize}
In this setting the functional \eqref{tikho} can be derived either
from a regularization point of view
\cite{girpog89,Wahba:splinesBook:1990} or from the theory of empirical
risk minimization (ERM) \cite{vapnik98}.  In the former, one observes
that, if the space $\hh_k$ is large enough, the minimization of the
empirical error is ill-posed, and in particular it responds in an
unstable manner to noise, or when the number of samples is low
Adding the squared norm stabilizes the problem. The latter point of view, starts from the
analysis of ERM showing that generalization to new samples can be
achieved if there is a tradeoff between fitting and
complexity\footnote{For example, a measure of complexity is the
  \emph{Vapnik–Chervonenkis dimension}
  \cite{Scholkopf:kernelsBook:2002}} of the estimator. The functional
\eqref{tikho} can be seen as an instance of such a trade-off.

The explicit form of the estimator is derived in two steps. First, one
can show that the minimizer of \eqref{tikho} can always be written as
a linear combination of the kernels centered at the training set
points,
$$
f_*(\boldx_*)=\sum_{i=1}^N k(\boldx_*,\boldx_i)c_i={\mathbf
  k}_{\boldx_*}^{\top}{\mathbf c},
$$
see for example \cite{micpon05,cadeto06}.  The above result is the
well known representer theorem originally proved in \cite{wahba70}
(see also \cite{Scholkopf01ageneralized} and \cite{repre} for recent
results and further references).  The explicit form of the
coefficients ${\mathbf c}=[c_1,\dots,c_N]^\top$ can be then derived by
substituting for $f_*(\boldx_*)$ in \eqref{tikho}.
\subsection{A Bayesian  Perspective}
A Gaussian process (GP) is a stochastic process with the important
characteristic that any finite number of random variables, taken from
a realization of the GP, follows a joint Gaussian distribution. A GP
is usually used as a prior distribution for functions
\cite{Rasmussen:book06}. If the function $f$ follows a Gaussian
process we write
\begin{align*}
f\sim \mathcal{GP}(m,k),
\end{align*}
where $m$ is the mean function and $k$ the covariance or kernel
function. The mean function and the covariance function completely
specify the Gaussian process.  In other words the above assumption
means that for any finite set $\boldX=\{\boldx_n\}_{n=1}^N$ if we let
$\fX=[f(\boldx_1), \dots, f(\boldx_N)]^\top$ then
\begin{align*}
\fX\sim\gauss(m(\boldX), \Kms),
\end{align*}
where $m(\boldX) = [m(\boldx_1), \dots, m(\boldx_N)]^\top$ and $\Kms$
is the kernel matrix.  In the following, unless otherwise stated, we
assume that the mean vector is zero.

From a \emph{Bayesian} point of view, the Gaussian process specifies
our prior beliefs about the properties of the function we are
modeling. Our beliefs are updated in the presence of data by means of
a \emph{likelihood function}, that relates our prior assumptions to
the actual observations. This leads to an updated distribution, the
\emph{posterior distribution}, that can be used, for example, for
predicting test cases.

In a regression context, the likelihood function is usually Gaussian
and expresses a linear relation between the observations and a given
model for the data that is corrupted with a zero mean Gaussian noise,
\begin{align*}
p(y|f,\boldx,\sigma^2)&=\gauss(f(\boldx), \sigma^2),
\end{align*}
where $\sigma^2$ corresponds to the variance of the noise. Noise is
assumed to be independent and identically distributed. In this way,
the likelihood function factorizes over data points, given the set of
inputs $\boldX$ and $\sigma^2$.  The posterior distribution can be
computed analytically. For a test input vector $\boldx_*$, given the
training data $\boldS = \{\boldX, \mathbf{Y}\}$, this posterior
distribution is given by,
\begin{align*}
  p(f(\boldx_*)|\boldS, \boldx_*, \bm{\phi})&=
  \gauss(f_*(\boldx_*), k_*(\boldx_*, \boldx_*)),
\end{align*}
where $\bm{\phi}$ denotes the set of parameters which include the variance of the noise, $\sigma^2$, and any parameters from the covariance function $k(\boldx,
\boldx')$. Here we have
\begin{align*}
  f_*(\boldx_*)&= \boldk^{\top}_{\boldx_*}(\Kms + \sigma^2\eye)^{-1}\mathbf{Y},\\
  k_*(\boldx_*, \boldx_*)&=k(\boldx_*,\boldx_*) -
  \boldk^{\top}_{\boldx_*}(\Kms + \sigma^2\eye)^{-1}\boldk_{\boldx_*}
\end{align*}
and finally we note that if we are interested into the distribution of
the noisy predictions, $p(y(\boldx_*) | \boldS, \boldx_*,
\bm{\phi})$, it is easy to see that we simply have to add $\sigma^2$
to the expression for the predictive variance (see
\cite{Rasmussen:book06}).

\begin{figure}[ht!]
\includegraphics[width=0.9\textwidth]{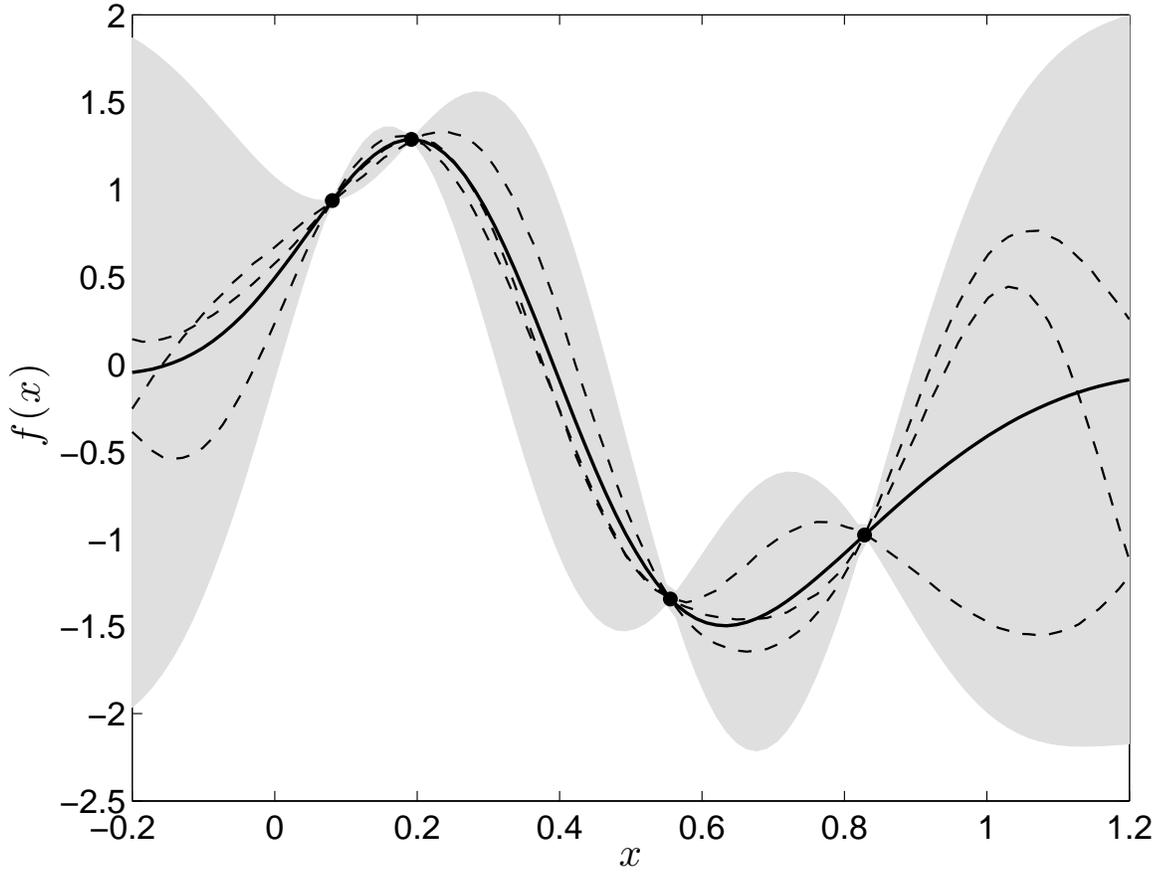}
\caption{Example of a predictive posterior distribution inferred with
  $N=4$. The solid line corresponds to the predictive mean, the shaded
  region corresponds to two standard deviations of the
  prediction. Dots are values of the output function $\mathbf{Y}$. We
  have also included some samples from the posterior distribution,
  shown as dashed lines.}
\label{fig:GP:1output}
\end{figure}

Figure \ref{fig:GP:1output} represents a posterior predictive
distribution for a data vector $\mathbf{Y}$ with $N=4$. Data points
are represented as dots in the figure. The solid line represents the
mean function predicted, $f_*(\boldx_*)$, while the shaded region
corresponds to two standard deviations away from the mean. This shaded
region is specified using the predicted covariance function,
$k_*(\boldx_*, \boldx_*)$. Notice how the uncertainty in the
prediction increases as we move away from the data points.

Equations for $f_*(\boldx_*)$ and $k_*(\boldx_*,\boldx_*)$ are
obtained under the assumption of a Gaussian likelihood, common in
regression setups. For non-Gaussian likelihoods, for example in
classification problems, closed form solutions are not longer
possible. In this case, one can resort to different approximations,
including the Laplace approximation and variational methods
\cite{Rasmussen:book06}.

\subsection{A Connection Between Bayesian and Regularization Point of  Views}

Connections between regularization theory and Gaussian process
prediction or Bayesian models for prediction have been pointed out
elsewhere
\cite{Poggio:Networks:1990,Wahba:splinesBook:1990,Rasmussen:book06}. Here
we just give a very brief sketch of the argument. We restrict
ourselves to finite dimensional RKHS. Under this assumption one can
show that every RKHS can be described in terms of a feature map
\cite{vapnik98}, that is a map $\Phi:\mathcal{X}\to \rone^p$, such
that
$$k(\boldx,\boldx')=\sum_{j=1}^p\Phi^j(\boldx)\Phi^j(\boldx').$$
In fact in this case one can show that functions in the RKHS with
kernel $k$ can be written as
$$f_w(\boldx)=\sum^p_{j=1} \boldw^j\Phi^j(\boldx)=\scal{\boldw}{\Phi(\boldx)},\quad \text{and} \quad\nor{f_\boldw}_k=\nor{\boldw}.$$
Then we can build a Gaussian process by assuming the coefficient
$w=w^1, \dots, w^p$ to be distributed according to a multivariate
Gaussian distribution. Roughly speaking, in this case the assumption
$f_*\sim \mathcal{GP}(0,k)$ becomes
$$
\boldw\sim {\mathcal N}(0, \mathbf{I}_p) \propto e^{-\nor{\boldw}^2}.
$$
As we noted
before if we assume a Gaussian likelihood we have
$$
P(\SY|\SX,f)={\mathcal N}(f(\SX),\sigma^2 \mathbf{I}_D)\propto e^{-\frac{1}{\sigma^2}\nor{f_w(\SX)-\SY}_n^2},
$$
where $f_\boldw(\SX)=(\scal{\boldw}{\Phi(\boldx_1)}, \dots,
\scal{\boldw}{\Phi(\boldx_n)})$ and
$\nor{f_w(\SX)-\SY}_n^2=\sum_{i=1}^n
(\scal{w}{\Phi(\boldx_i)}-y_i)^2$.  Then the posterior distribution is
proportional to
$$
e^{-(\frac{1}{\sigma^2}\nor{f_\boldw(\SX)-\SY}_n^2+\nor{\boldw}^2)},
$$
and we see that a maximum a posteriori estimate will in turn give the
minimization problem defining Tikhonov regularization \cite{tikars77}, where the
regularization parameter is now related to the noise variance.

We note that in regularization the squared error is often replaced by
a more general error term $\frac{1}{N}\sum_{i=1}^N
\ell(f(\boldx_i),y_i)$.  In a regularization perspective, the {\em
  loss function} $\ell:\rone\times\rone \to\rone^+$ measure the error
we incur when predicting $f(\boldx)$ in place of $y$.  The choice of
the loss function is \emph{problem dependent}. Often used examples are
the square loss, the logistic loss or the hinge loss used in support
vector machines (see \cite{Scholkopf:kernelsBook:2002}).

The choice of a loss function in a regularization setting can be
contrasted to the choice of the likelihood in a Bayesian setting. In
this context, the likelihood function models how the observations
deviate from the assumed {\em true} model in the generative
process. The notion of a loss function is philosophically
different. It represents the cost we pay for making errors. In
Bayesian modeling decision making is separated from inference. In the
inference stage the posterior distributions are computed evaluating
the uncertainty in the model. The loss function appears only at the
second stage of the analysis, known as the \emph{decision} stage, and
weighs how incorrect decisions are penalized given the current
uncertainty. However, whilst the two notions are philosophically very
different, we can see that, due to the formulation of the frameworks,
the loss function and the log likelihood provide the same role
mathematically.

The discussion in the previous sections shows that the notion of a
kernel plays a crucial role in statistical modeling both in the
Bayesian perspective (as the covariance function of a GP) and the
regularization perspective (as a reproducing kernel). Indeed, for
scalar valued problems there is a rich literature on the design of
kernels (see for example
\cite{Scholkopf:kernelsBook:2002,tacri04,Rasmussen:book06} and
references therein). In the next sections we show how the concept of a
kernel can be used in multi-output learning problems.  Before doing
that, we describe how the concepts of RKHSs and GPs translate to the
setting of vector valued learning.

\section{ Learning Multiple Outputs with Kernels Methods }
\label{chap:VecLearn}
In this chapter we discuss the basic setting for learning vector
valued functions and related problems (multiclass, multilabel) and
then describe how the concept of kernels (reproducing kernels and
covariance function for GP) translate to this setting.

\subsection{Multi-output Learning}\label{Sec:VecLearn}
The problem we are interested in is that of learning an unknown
functional relationship $f$ between an input space $\cal X$, for
example ${\cal X}=\rone^p$, and an output space $\rone^D$.  In the
following we will see that the problem can be tackled either assuming
that $\boldf$ belongs to reproducing kernel Hilbert space of vector
valued functions or assuming that $\boldf$ is drawn from a vector
valued Gaussian process.  Before doing this we describe several
related settings all falling under the framework of multi-output
learning.

The natural extension of the traditional (scalar) supervised learning
problem is the one we discussed in the introduction, when the data are
pairs $S=(\SX, \SY)=(\boldx_1,y_1),\dots, (\boldx_N,y_N)$. For example
this is the typical setting for problems such as motion/velocity
fields estimation. A special case is that of multi-category
classification problem or multi-label problems, where if we have $D$
classes each input point can be associated to a (binary) coding vector
where, for example $1$ stands for presence ($0$ for absence) of a
class instance.The simplest example is the so called {\em one vs all}
approach to multiclass classification which, if we have $\{1, \dots,
D\}$ classes, amounts to the coding $i\to {\mathbf e}_i$, where
$({\mathbf e}_i)$ is the canonical basis of $\rone^D$.

A more general situation is that where different outputs might have
different training set cardinalities, different input points or in the
extreme case even different input spaces. More formally, in this case
we have a training set $S_d=(\SX_d,
\SY_d)=(\boldx_{d,1},y_{d,1}),\dots, (\boldx_{d,N_d},y_{d,N_d})$ for
each component $f_d$, with $d=1,\dots, D$, where the number of data
associated with each output, $(N_d)$ might be different and the input
for a component might belong to different input space $(\X_d)$.

The terminology used in machine learning often does not distinguish
the different settings above and the term multitask learning is often
used.  In this paper we use the term multi-output learning or vector
valued learning to define the general class of problems and use the
term multi-task for the case where each component has different
inputs. Indeed in this very general situation each component can be
thought of as a distinct task possibly related to other tasks
(components).  In the geostatistics literature, if each output has the
same set of inputs the model is called \emph{isotopic} and
\emph{heterotopic} if each output to be associated with a different
set of inputs \cite{Wackernagel:book03}.  Heterotopic data is further
classified into \emph{entirely heterotopic data}, where the variables
have no sample locations in common, and \emph{partially heterotopic
  data}, where the variables share some sample locations. In machine
learning, the partially heterotopic case is sometimes referred to as
\emph{asymmetric multitask learning}
\cite{Xue:MTLDirichlet:2007,Chai:learningcurvesMTGP:2009}.

The notation in the multitask learning scenario (heterotopic case) is
a bit more involved.  To simplify the notation we assume that the
number of data for each output is the same.  Moreover, for the sake of
simplicity sometimes we restrict the presentation to the isotopic
setting, though the models can usually readily be extended to the more
general setting.  We will use the notation $\boldX$ to indicate the
collection of all the training input points, $\{\SX_j\}_{j=1}^N$, and
${\boldS}$ to denote the collection of all the training data. Also we
will use the notation $\FX$ to indicate a vector valued function
evaluated at different training points. This notation has slightly
different meaning depending on the way the input points are
sampled. If the input to all the components are the same then
$\boldX=\boldx_1, \dots, \boldx_N$ and $\FX=f_1(\boldx_1), \dots,
f_D(\boldx_N)$.  If the input for the different components are
different then $\boldX=\{\boldX_d\}_{d=1}^D= \boldX_1, \dots,
\boldX_D$, where $\boldX_d=\{\boldx_{d,n}\}_{n=1}^N$ and $\FX=
(f_1(\boldx_{1,1}), \dots, f_1(\boldx_{1,N})), \dots,
(f_D(\boldx_{D,1}), \dots, f_D(\boldx_{D,N}))$.

\subsection{Reproducing Kernel for Vector Valued Function}\label{subsection:reg:mol:def}

The definition of RKHS for vector valued functions parallels the one
in the scalar, with the main difference that the reproducing kernel is
now {\em matrix} valued, see for example \cite{micpon05,cadeto06} .  A
reproducing kernel is a symmetric function $\vk:\mathcal{X}\times
\mathcal{X}\rightarrow \rone^{D\times D}$, such that for any
$\boldx,\boldx'$ $\vk(\boldx,\boldx')$ is a positive semi-definite
{\em matrix}.  A vector valued RKHS is a Hilbert space $\mathcal{H}$
of functions $\boldf:\mathcal{X}\to \R^D$, such that for very
$\boldc\in \rone^D$, and $\mathbf{x}\in \mathcal{X}$, $\vk(\mathbf{x},
\mathbf{x}')\boldc$, as a function of $\mathbf{x}'$ belongs to
$\mathcal{H}$ and moreover $\vk$ has the reproducing property
 $$\langle \boldf,\vk(\cdot, \mathbf{x})\boldc \rangle_{\vk} =\boldf(\mathbf{x})^\top\boldc,$$
 where $\langle \cdot,\cdot \rangle_{\vk}$ is the inner product in
 ${\mathcal{H}}$.

Again, the choice of the kernel corresponds to the choice of the
representation (parameterization) for the function of interest. In
fact any function in the RKHS is in the closure of the set of  linear combinations
$$
\boldf(\boldx)=\sum_{i=1}^p \vk(\boldx_i,\boldx)\boldc_j,~~~~\boldc_j\in \rone^D,
$$
where we note that in the above equation each term
$\vk(\boldx_i,\boldx)$ is a matrix acting on a vector $\boldc_j$.  The
norm in the RKHS typically provides a measure of the complexity of a
function and this will be the subject of the next sections.

Note that the definition of vector valued RKHS can be described in a
component-wise fashion in the following sense.  The kernel $\vk$ can
be described by a scalar kernel $R$ acting jointly on input examples
and task indices, that is
\begin{equation}\label{CompWiseKer}
(\vk(\boldx,\boldx'))_{d,d'}=R((\boldx,d),(\boldx',d')),
\end{equation}
where $R$ is a scalar reproducing kernel on the space
$\mathcal{X}\times\{1,\dots,D\}$. This latter point of view is useful
while dealing with multitask learning, see \cite{Evgeniou:multitask05}
for a discussion.

Provided with the above concepts we can follow a regularization
approach to define an estimator by minimizing the regularized
empirical error \eqref{tikho}, which in this case can be written as
\begin{align}\label{tikhofuncvec}
\sum_{j=1}^D\frac{1}{N}\sum_{i=1}^N (f_j(\boldx_i)-y_{j,i})^2+\la \nor{\boldf}^2_\vk,
\end{align}
where $\boldf=(f_1, \dots, f_D)$.
Once again the solution is given by the representer theorem   \cite{micpon05}
$$
\boldf(\boldx)=\sum_{i=1}^N\vk(\boldx_i,\boldx)\boldc_i,
$$
and the coefficient satisfies the linear system
\begin{equation}\label{vectortikho}
\boldcv=(\Kmv+\la N\eye )^{-1}\boldyv,
\end{equation}
where $\boldcv, \boldyv$ are $ND$ vectors obtained concatenating the
coefficients and the output vectors, and $\Kmv$ is an $ND\times ND$
with entries $(\vk(\boldx_i,\boldx_j))_{d,d'}$, for $i,j=1,\dots, N$
and $d,d'=1,\dots,D$ (see for example \cite{micpon05}).  More
explicitly
\begin{align}\label{bigmat}
\Kmv &=
\begin{bmatrix}
\Kmsd{1}{1} &  \cdots & \Kmsd{1}{D}\\
\Kmsd{2}{1} &  \cdots & \Kmsd{2}{D}\\
   \vdots    & \cdots &     \vdots             \\
\Kmsd{D}{1}& \cdots & \Kmsd{D}{D}\\
\end{bmatrix}
\end{align}
where each block $\Kmsd{i}{j}$ is an $N$ by $N$ matrix (here we make
the simplifying assumption that each output has same number of
training data). Note that given a new point $\boldx_*$ the
corresponding prediction is given by
$$
\boldf(\boldx_*)=\boldK_{\boldx_*}^\top \boldcv,
$$
where $\boldK_{\boldx_*}\in \rone^{D\times ND}$ has entries $(\vk(\boldx_*, \boldx_j))_{d,d'}$ for $j=1,\dots, N$ and
$d,d'=1, \ldots, D$.

\subsection{Gaussian Processes for Vector Valued Functions}

Gaussian process methods for modeling vector-valued functions follow
the same approach as in the single output case. Recall that a Gaussian
process is defined as a collection of random variables, such that any
finite number of them follows a joint Gaussian distribution. In the
single output case, the random variables are associated to a single
process $f$ evaluated at different values of $\boldx$ while in the
multiple output case, the random variables are associated to different
processes $\{f_d\}_{d=1}^D$, evaluated at different values of $\boldx$
\cite{Cressie:spatialdataBook:1993,Goovaerts:book97,verHoef:convolution98}.

The vector-valued function $\mathbf{f}$ is assumed to follow a Gaussian process
\begin{align}\label{equation:def:general:gp:multi:output}
\mathbf{f}\sim \mathcal{GP}(\mathbf{m},\vk),
\end{align}
where $\mathbf{m}\in\rone^D$ is a vector which components are the mean
functions $\{m_d(\mathbf{x})\}_{d=1}^D$ of each output and $\vk$ is a
positive \emph{matrix} valued function as in section
\ref{subsection:reg:mol:def}. The entries
$(\vk(\boldx,\boldx'))_{d,d'}$ in the matrix $\vk(\boldx, \boldx')$
correspond to the covariances between the outputs $f_d(\boldx)$ and
$f_{d'}(\boldx')$ and express the degree of correlation or similarity
between them.

For a set of inputs $\boldX$, the prior distribution over the vector $\FX$ is given by
\begin{align*}
\FX\sim \gauss(\mathbf{m}(\boldX), \vk(\boldX,\boldX)),
\end{align*}
where $\mathbf{m}(\boldX)$ is a vector that concatenates the mean
vectors associated to the outputs and the covariance matrix $\Kmv$ is
the block partitioned matrix in \eqref{bigmat}.  Without loss of
generality, we assume the mean vector to be zero.

In a regression context, the likelihood function for the outputs is
often taken to be Gaussian distribution, so that
\begin{align*}
p(\boldy|\boldf,\boldx,\Sigma) &= \gauss(\boldf(\boldx), \Sigma),
\end{align*}
where $\Sigma\in\rone^{D\times D}$ is a diagonal matrix with elements\footnote{This relation derives from $y_d(\boldx)=f_d(\boldx) + \epsilon_d(\boldx)$, for each $d$,
where $\{\epsilon_d(\boldx)\}_{d=1}^D$ are independent white Gaussian noise processes with variance $\sigma^2_d$.
} $\{\sigma^2_{d}\}_{d=1}^D$.

For a Gaussian likelihood, the predictive distribution and the
marginal likelihood can be derived analytically.  The predictive
distribution for a new vector $\boldx_{*}$ is \cite{Rasmussen:book06}
\begin{align}\label{eq:predictive:cmoc}
p(\boldf(\boldx_*)|{\mathbf S},\boldf,\boldx_*,\bm{\phi})&=
\mathcal{N}\left(\boldf_*(\boldx_*),\vk_*(\boldx_*, \boldx_*)\right),
\end{align}
with
\begin{align*}
\boldf_*(\boldx_*)&=\boldK_{\boldx_*}^\top(\Kmv+\bm{\Sigma})^{-1}\overline{\mathbf{y}},\\
\vk_*(\boldx_*, \boldx_*)&= \vk(\boldx_*, \boldx_*)-\boldK_{\boldx_*}(\Kmv+\bm{\Sigma})^{-1}
\boldK^{\top}_{\boldx_*},
\end{align*}
where $\bm{\Sigma}=\Sigma\otimes\mathbf{I}_N$, $\boldK_{\boldx_*}\in
\rone^{D\times ND}$ has entries $(\vk(\boldx_*, \boldx_j))_{d,d'}$ for
$j=1,\dots, N$ and $d,d'=1, \ldots, D$, and $\bm{\phi}$ denotes a
possible set of hyperparameters of the covariance function
$\vk(\boldx, \boldx')$ used to compute $\Kmv$ and the variances of the
noise for each output $\{\sigma^2_{d}\}_{d=1}^D$.  Again we note that
if we are interested into the distribution of the noisy predictions it
is easy to see that we simply have to add $\bm{P\Sigma}$ to the
expression of the prediction variance. The above expression for the
mean prediction coincides again with the prediction of the estimator
derived in the regularization framework.

In the following chapters we describe several possible choices of
kernels (covariance function) for multi-output problems.  We start in
the next chapter with kernel functions that clearly separate the
contributions of input and output.  We will see later alternative ways
to construct kernel functions that interleave both contributions in a
non trivial way.

\section{Separable Kernels and Sum of Separable Kernels}\label{SepKer}
In this chapter we review a special class of multi-output kernel
functions that can be formulated as a sum of products between a kernel function for
the input space alone, and a kernel function that encodes the
interactions among the outputs.  We refer to this type of
multi-output kernel functions as {\em separable kernels} and {\em sum of separable kernels} (SoS kernels).

We consider a class of kernels  of the form
$$
(\vk(\boldx,\boldx'))_{d,d'}=k(\boldx,\boldx')k_T(d,d'),
$$
where $k, k_T$ are scalar kernels on $\mathcal{X}\times {\mathcal X}$ and $\{1, \dots, D\}\times \{1, \dots, D\}$.

Equivalently one can consider the matrix expression
\begin{equation}\label{SplitKernel}
\vk(\boldx,\boldx')=k(\boldx,\boldx')\boldB,
\end{equation}
where $\boldB$ is a $D\times D$ symmetric and positive semi-definite matrix. We call this class of kernels separable
since, comparing to \eqref{CompWiseKer}, we see that the contribution of input and output
is decoupled.

In the same spirit a more general class of kernels is given by
$$
\vk(\boldx,\boldx')=\sum_{q=1}^Qk_q(\boldx,\boldx')\boldB_q.
$$
For this class of kernels, the kernel matrix associated
to a data set $\boldX$ has a simpler form and can be written as

\begin{align}\label{eq:matricesLMC}
\Kmv &= \sum_{q=1}^Q    \boldB_q \otimes k_q(\boldX, \boldX),
\end{align}
where  $\otimes$ represents the Kronecker product between matrices. We call this class of kernels sum of separable
kernels (SoS kernels).

The simplest example of separable kernel is given by setting
$k_T(d,d')=\delta_{d,d'}$, where $\delta_{d,d'}$ is the Kronecker
delta.  In this case $\boldB=\Id_N$, that is all the outputs are
treated as being unrelated. In this case the kernel matrix $\Kmv$,
associated to some set of data $\boldX$, becomes block diagonal. Since
the off diagonal terms encode output relatedness.  We can see that the
matrix $\boldB$ encodes dependencies among the outputs.

The key question is how to choose the scalar kernels $\{k_q\}_{q=1}^Q$
and especially how to design, or learn, the matrices
$\{\boldB_q\}_{q=1}^Q$.  This is the subject we discuss in the next
few sections.  We will see that one can approach the problem from a
regularization point of view, where kernels will be defined by the
choice of suitable regularizers, or, from a Bayesian point of view,
constructing covariance functions from explicit generative models for
the different output components. As it turns out these two points of view
are equivalent and allow for two different interpretations of the same
class of models.
\subsection{Kernels and Regularizers}

In this section we largely follow the results in
\cite{micchelli04kernels,micpon05,evgeniou05multitask} and
\cite{BRBV10}.  A possible way to design multi-output kernels of the
form \eqref{SplitKernel} is given by the following result.  If $\vk$
is given by \eqref{SplitKernel} then is possible to prove that the
norm of a function in the corresponding RKHS can be written as
\beeq{eq:prop1}{ \nor{\boldf}_\vk^2=\sum_{d,d'=1}^D \boldB_{d,
    d'}^\dagger\scal{f_d}{f_{d'}}_k, } where $\boldB^\dagger$ is the
pseudoinverse of $\boldB$ and $\boldf=(f_1, \dots, f_D)$.  The above
expression gives another way to see why the matrix $\boldB$ encodes
the relation among the components.  In fact, we can interpret the
right hand side in the above expression as a regularizer inducing
specific coupling among different tasks $\scal{f_t}{f_{t'}}_k$ with
different weights given by $\boldB_{d,d'}^\dagger$.  This result says
that any such regularizer induces a kernel of the form
\eqref{SplitKernel}.  We illustrate the above idea with a few
examples.

\paragraph{Mixed Effect Regularizer}

 Consider the regularizer given by
\beeq{eq:reg_simple}{ R(\boldf) = A_\omega \left(C_\omega
    \sum_{\ell=1}^D \nor{f_\ell}^2_k + \omega D \sum_{\ell=1}^D
    \nor{f_{\ell}-\frac{1}{D}\sum_{q=1}^D f_q}^2_k \right) } where $
A_\omega=\frac{1}{2(1-\omega)(1-\omega+\omega D)} $ and $
C_\omega=(2-2\omega+\omega D).  $ The above regularizer is composed of
two terms: the first is a standard regularization term on the norm of
each component of the estimator; the second forces each $f_\ell$ to be
close to the mean estimator across the components, $\overline{f} =
\frac{1}{D} \sum_{q=1}^D f_q$.  The corresponding kernel imposes a
common similarity structure between all the output components and the
strength of the similarity is controlled by a parameter $\omega$,
\beeq{mixed}{ \vk_\omega(\boldx,\boldx') = k(\boldx,\boldx')(\omega \1
  + (1-\omega) \Id_D) } where $\1$ is the $D \times D$ matrix whose
entries are all equal to $1$, and $k$ is a scalar kernel on the input
space $\X$. Setting $\omega = 0$ corresponds to treating all
components independently and the possible similarity among them is not
exploited. Conversely, $\omega = 1$ is equivalent to assuming that all
components are identical and are explained by the same function. By
tuning the parameter $\omega$ the above kernel interpolates between
this two opposites cases. We note that from a Bayesian perspective $B$
is a correlation matrix with all the off-diagonals equal to $\omega$,
which means that the output of the Gaussian process are exchangeable.

\paragraph{Cluster Based Regularizer.} Another example of regularizer,
proposed in \cite{Evgeniou:multitask05}, is based on the idea of
grouping the components into $r$ clusters and enforcing the components
in each cluster to be similar. Following \cite{jacob08clusteredmtl},
let us define the matrix $\boldE$ as the $D \times r$ matrix, where
$r$ is the number of clusters, such that $\boldE_{\ell,c} = 1$ if the
component $l$ belongs to cluster $c$ and $0$ otherwise. Then we can
compute the $D \times D$ matrix $\boldM =
\boldE(\boldE^\top\boldE)^{-1}\boldE^\top$ such that $\boldM_{\ell,q}
= \frac{1}{m_c}$ if components $l$ and $q$ belong to the same cluster
$c$, and $m_c$ is its cardinality, $\boldM_{\ell,q} = 0$
otherwise. Furthermore let $I(c)$ be the index set of the components
that belong to cluster $c$. Then we can consider the following
regularizer that forces components belonging to the same cluster to be
close to each other: \beeq{eq:clustered}{ R(\boldf) = \epsilon_1
  \sum_{c=1}^r \sum_{\ell \in I(c)} \nor{f_\ell - \overline{f}_c}^2_k
  + \epsilon_2 \sum_{c=1}^r m_c \nor{\overline{f}_c}^2_k, } where
$\overline{f}_c$ is the mean of the components in cluster $c$ and
$\epsilon_1, \epsilon_2$ are parameters balancing the two terms.
Straightforward calculations show that the previous regularizer can be
rewritten as $R(\boldf) =
\sum_{\ell,q}\boldG_{\ell,q}\scal{f_\ell}{f_q}_k$, where

\begin{equation}\label{graphreg}
\boldG_{\ell, q} = \epsilon_1 \delta_{lq} + (\epsilon_2 - \epsilon_1) \boldM_{\ell, q}.
\end{equation}
Therefore the corresponding matrix valued kernel is
$\vk(\boldx,\boldx') = k(\boldx,\boldx')\boldG^\dagger$.

\paragraph{Graph Regularizer.} Following
\cite{micchelli04kernels,sheldon}, we can define a regularizer that,
in addition to a standard regularization on the single components,
forces stronger or weaker similarity between them through a given $D
\times D$ positive weight matrix $\boldM$,
\beeq{eq:reg}{
R(\boldf) = \frac{1}{2} \sum_{\ell,q=1}^D \nor{f_\ell - f_q}^2_k \boldM_{\ell q} + \sum_{\ell=1}^D \nor{f_\ell}^2_k \boldM_{\ell, \ell}.
}
The regularizer $J(f)$ can be rewritten as:
\mfni
\sum_{\ell,q=1}^D \lp \nor{f_\ell}^2_k \boldM_{\ell, q} - \scal{f_\ell}{f_q}_k\boldM_{\ell, q} \rp + \sum_{\ell=1}^D \nor{f_\ell}^2_k \boldM_{\ell,\ell} & = &\nonumber\\
\sum_{\ell=1}^D \nor{f_\ell}^2_k \sum_{q=1}^D (1 + \delta_{\ell, q})\boldM_{\ell, q} - \sum_{\ell,q=1}^D \scal{f_\ell}{f_q}_k\boldM_{\ell, q} & = &\nonumber\\
\sum_{\ell,q=1}^D \scal{f_\ell}{f_q}_k \boldL_{\ell, q} & & \label{eq:reg2}
\mfnf
where $\boldL = \boldD - \boldM$, with $\boldD_{\ell, q} = \delta_{\ell, q} \lp \sum_{h=1}^D \boldM_{\ell, h} +  \boldM_{\ell, q} \rp$.
Therefore the resulting kernel will be $\vk(\boldx,\boldx') = k(\boldx,\boldx') \boldL^\dagger$, with $k(\boldx,\boldx')$
 a scalar kernel to be chosen according to the problem at hand.

 In the next section we will see how models related to those described
 above can be derived from suitable generative models.

\subsection{Coregionalization Models}

The use of probabilistic models and Gaussian processes for
multi-output learning was pioneered and largely developed in the
context of geostatistics, where prediction over vector-valued output
data is known as \emph{cokriging}.  Geostatistical approaches to
multivariate modelling are mostly formulated around the ``linear model
of coregionalization'' (LMC)
\cite{Journel:miningBook78,Goovaerts:book97}, that can be considered
as a generative approach for developing valid covariance
functions. Covariance functions obtained under the LMC assumption
follow the form of a sum of separable kernels. We will start considering this model and then
discuss how several models recently proposed in the machine learning
literature are special cases of the LMC.

\subsubsection{The Linear Model of Coregionalization}\label{section:lmc}

In the linear model of coregionalization, the outputs are expressed as
linear combinations of independent random functions. This is done in a
way that ensures that the resulting covariance function (expressed
jointly over all the outputs and the inputs) is a valid positive
semidefinite function.  Consider a set of $D$ outputs
$\{f_d(\mathbf{x})\}_{d=1}^D$ with $\boldx\in\rone^p$. In the LMC,
each component $f_d$ is expressed as \cite{Journel:miningBook78}
\begin{align}\notag
f_d(\boldx) &= \sum_{q =1}^Qa_{d,q}u_{q}(\boldx),
\end{align}
where the latent functions $u_{q}(\boldx)$, have mean zero and
covariance $\cov[u_{q}(\boldx),u_{q'}(\boldx')]=k_{q}(\boldx,\boldx')$
if $q=q'$, and $a_{d,q}$ are scalar coefficients.  The processes
$\{u_{q}(\boldx)\}_{q=1}^Q$ are independent for $q\neq q'$. The
independence assumption can be relaxed and such relaxation is
presented as an extension in section
\ref{separable:kernels:extensions}. Some of the basic processes
$u_{q}(\boldx)$ and $u_{q'}(\boldx')$ can have the same covariance
$k_{q}(\boldx,\boldx')$, while remaining independent.

A similar expression for $\{f_d(\mathbf{x})\}_{d=1}^D$ can be written
grouping the functions $u_{q}(\boldx)$ which share the same covariance
\cite{Journel:miningBook78,Goovaerts:book97}
\begin{align}\label{eq:lmc}
f_d(\boldx) &= \sum_{q =1}^Q\sum_{i=1}^{R_{q}}a_{d,q}^{i}u_q^{i}(\boldx),
\end{align}
where the functions $u_q^{i}(\boldx)$, with $q=1,\ldots, Q$ and
$i=1,\ldots, R_q$, have mean equal to zero and covariance
$\cov[u_q^{i}(\boldx),u_{q'}^{i'}(\boldx')]=k_{q}(\boldx,\boldx')$ if
$i=i'$ and $q=q'$. Expression \eqref{eq:lmc} means that there are $Q$
groups of functions $u_q^{i}(\boldx)$ and that the functions
$u_q^{i}(\boldx)$ within each group share the same covariance, but are
independent.  The cross covariance between any two functions
$f_d(\boldx)$ and $f_{d'}(\boldx)$ is given in terms of the covariance
functions for $u_{q}^{i}(\boldx)$
\begin{align}\notag
\cov[f_d(\boldx),f_{d'}(\boldx')]&=\sum_{q=1}^Q\sum_{q'=1}^Q\sum_{i=1}^{R_q}\sum_{i'=1}^{R_q}a_{d,q}^ia_{d',q'}^{i'}
\cov[u_q^{i}(\boldx),u_{q'}^{i'}(\boldx')].
\end{align}
The covariance $\cov[f_d(\boldx),f_{d'}(\boldx')]$ is given by
$(\vk(\boldx, \boldx'))_{d,d'}$. Due to the independence of the
functions $u_{q}^{i}(\boldx)$, the above expression reduces to
\begin{align}\label{eq:lmc:fullCov}
(\vk(\boldx, \boldx'))_{d,d'}&=\sum_{q=1}^Q\sum_{i=1}^{R_q}a_{d,q}^ia_{d',q}^{i}k_{q}(\boldx, \boldx')
=\sum_{q=1}^Q b_{d,d'}^qk_{q}(\boldx, \boldx'),
\end{align}
with $b_{d,d'}^q=\sum_{i=1}^{R_q}a_{d,q}^ia_{d',q}^{i}$. The kernel
$\vk(\boldx, \boldx')$ can now be expressed as
\begin{align}
\vk(\boldx, \boldx') &= \sum_{q=1}^Q\boldB_qk_q(\boldx, \boldx'),\label{eq:multi-out:lmc:fullCov:matrixForm}
\end{align}
where each $\boldB_q\in\rone^{D\times D}$ is known as a
\emph{coregionalization matrix}. The elements of each $\boldB_q$ are
the coefficients $b_{d,d'}^q$ appearing in equation
\eqref{eq:lmc:fullCov}. The rank for each matrix $\boldB_q$ is
determined by the number of latent functions that share the same
covariance function $k_q(\boldx, \boldx')$, that is, by the
coefficient $R_q$.

Equation \eqref{eq:lmc} can be interpreted as a nested structure
\cite{Wackernagel:book03} in which the outputs $f_d(\boldx)$ are first
expressed as a linear combination of spatially uncorrelated processes
$f_d(\boldx) =\sum_{q=1}^Qf_d^q(\boldx),$ with $\ex[f_d^q(\boldx)]=0$
and $\cov[f_d^q(\boldx),f_{d'}^{q'}(\boldx')]=b_{d,d'}^qk_q(\boldx,
\boldx')$ if $q=q'$, otherwise it is equal to zero.  At the same time,
each process $f_d^q(\boldx)$ can be represented as a set of
uncorrelated functions weighted by the coefficients $a_{d,q}^i$,
$f_d^q(\boldx) =\sum_{i=1}^{R_q}a_{d,q}^i u_q^i(\boldx)$ where again,
the covariance function for $u^i_q(\boldx)$ is $k_q(\boldx,\boldx')$.

Therefore, starting from a generative model for the outputs, the
linear model of coregionalization leads to a sum of separable kernels that
represents the covariance function as the sum of the products of two
covariance functions, one that models the dependence between the
outputs, independently of the input vector $\boldx$ (the
coregionalization matrix $\boldB_q$), and one that models the input
dependence, independently of the particular set of functions
$\{f_d(\boldx)\}$ (the covariance function $k_q(\boldx, \boldx')$).
The covariance matrix for $\FX$ is given by \eqref{eq:matricesLMC}.
\subsubsection{Intrinsic Coregionalization Model}

A simplified version of the LMC, known as the intrinsic
coregionalization model (ICM) (see \cite{Goovaerts:book97}), assumes
that the elements $b_{d,d'}^q$ of the coregionalization matrix
$\boldB_q$ can be written as $b_{d,d'}^q=\upsilon_{d,d'}b_q$, for some
suitable coefficients $\upsilon_{d,d'}$.
With this form for
$b_{d,d'}^q$, we have
\begin{align*}
  \cov[f_d(\boldx),f_{d'}(\boldx')]&=\sum_{q=1}^Q\upsilon_{d,d'}b_qk_{q}(\boldx,
  \boldx'),
  =\upsilon_{d,d'}\sum_{q=1}^Q b_qk_{q}(\boldx,\boldx')\\
  &= \upsilon_{d,d'}k(\boldx,\boldx'),
\end{align*}
where $k(\boldx,\boldx')=\sum_{q=1}^Q b_qk_{q}(\boldx,\boldx')$.  The
above expression can be seen as a particular case of the kernel
function obtained from the linear model of coregionalization, with
$Q=1$. In this case, the coefficients
$\upsilon_{d,d'}=\sum_{i=1}^{R_1}a_{d,1}^ia_{d',1}^{i}=b_{d,d'}^1$,
and the kernel matrix for multiple outputs becomes
$\boldK(\boldx,\boldx')= k(\boldx,\boldx')\boldB$ as in
\eqref{SplitKernel}.

The kernel matrix corresponding to a dataset $\boldX$  takes the form
\begin{align}\label{eq:matricesICM}
\Kmv &=  \boldB  \otimes \Kms.
\end{align}
One can see that the intrinsic coregionalization model corresponds to
the special separable kernel often used in the context of
regularization.
Notice that the value of $R_1$ for the coefficients
$\upsilon_{d,d'}=\sum_{i=1}^{R_1}a_{d,1}^ia_{d',1}^{i}=b_{d,d'}^1$,
determines the rank of the matrix $\boldB$.

As pointed out by \cite{Goovaerts:book97}, the ICM is much more
restrictive than the LMC since it assumes that each basic covariance
$k_q(\boldx, \boldx')$ contributes equally to the construction of the
autocovariances and cross covariances for the outputs. However, the
computations required for the corresponding inference are greatly
simplified, essentially because of the properties of the Kronecker
product. This latter point is discussed in detail in
Section~\ref{parEstimation}.

It can be shown that if the outputs are considered to be noise-free,
prediction using the intrinsic coregionalization model under an
isotopic data case is equivalent to independent prediction over each
output \cite{Helterbrand:universalCR94}. This circumstance is also
known as autokrigeability \cite{Wackernagel:book03}.

\begin{figure}[ht!]
\centering
\includegraphics[width=0.8\textwidth]{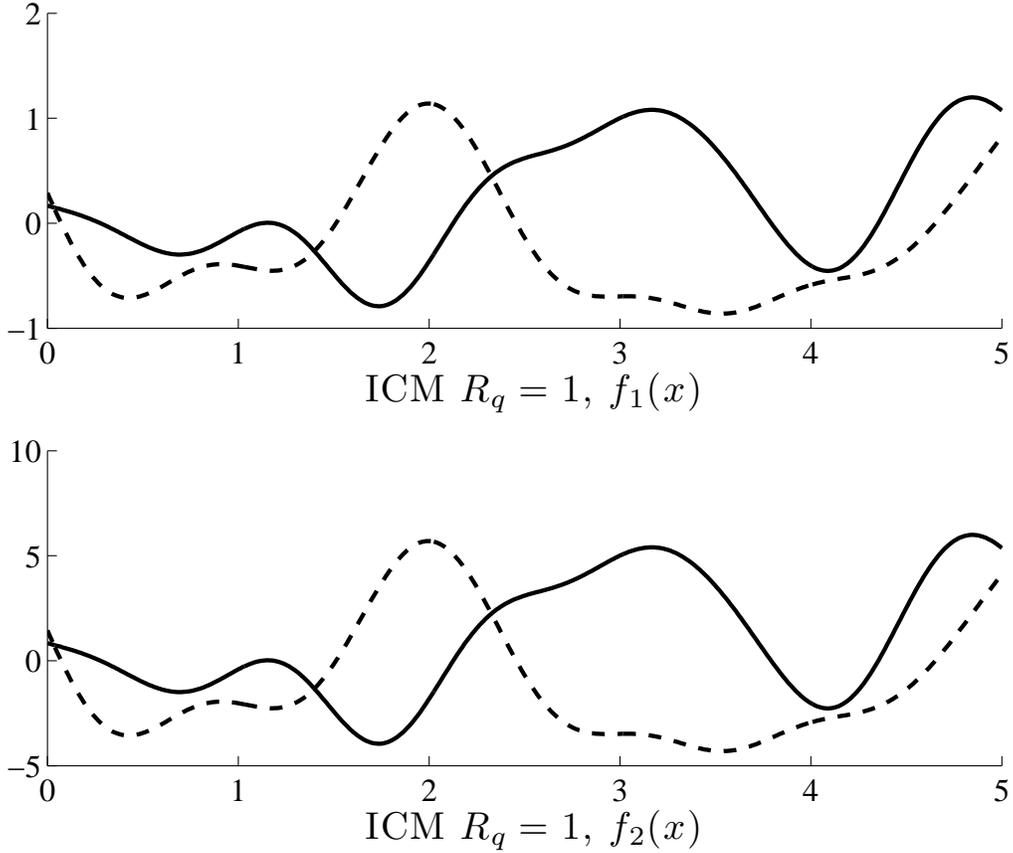}
\caption{Two samples from the intrinsic coregionalization model with rank one, this is $R_q=1$. Solid lines represent one
of the samples, and dashed lines represent the other sample. Samples are identical except for scale.}
\label{fig:ICM:Rank1}
\end{figure}

\subsubsection{Comparison Between ICM and LMC}

We have seen before that the intrinsic coregionalization model is a
particular case of the linear model of coregionalization for $Q=1$
(with $R_q\ne 1$) in equation \ref{eq:lmc:fullCov}. Here we contrast
these two models.  Note that a different particular case of the linear
model of coregionalization is assuming $R_q =1$ (with $Q\ne 1$). This
model, known in the machine learning literature as the semiparametric
latent factor model (SLFM) \cite{Teh:semiparametric05}, will be
introduced in the next subsection.

To compare the two models we have sampled from a multi-output Gaussian
process with two outputs ($D=2$), a one-dimensional input space ($x\in
\rone$) and a LMC with different values for $R_q$ and $Q$. As basic
kernels $k_q(\boldx, \boldx')$ we have used the \rbfKernelLong\
(\rbfKernel) kernel given as \cite{Rasmussen:book06},
\begin{align*}
k_q(\boldx, \boldx') & = \exp\left(-\frac{\lVert \boldx - \boldx' \rVert^2}{\ell_q^2}\right),
\end{align*}
where $\lVert\cdot\rVert$ represents the Euclidian norm and $\ell_q$
is known as the characteristic length-scale. The exponentiated
quadratic is variously referred to as the Gaussian, the radial basis
function or the squared exponential kernel.

\begin{figure}[ht!]
\centering
\includegraphics[width=0.8\textwidth]{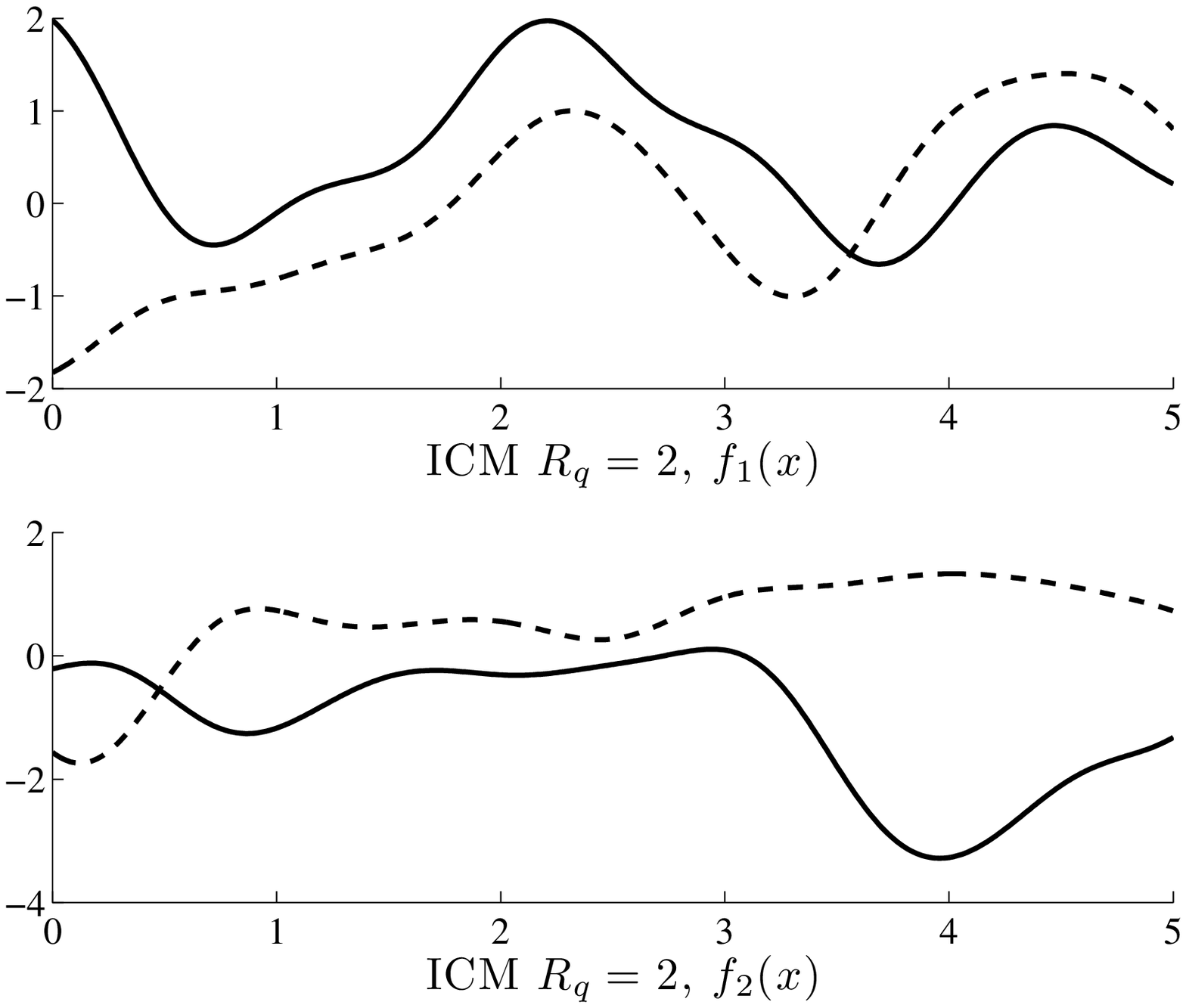}
\caption{Two samples from the intrinsic coregionalization model with
  rank two, $R_q=2$. Solid lines and dashed lines represent different samples. Although samples from different outputs
have the same length-scale, they look different and are not simply scaled versions of one another.}
\label{fig:ICM:Rank2}
\end{figure}
Figure \ref{fig:ICM:Rank1} shows samples from the intrinsic
coregionalization model for $R_q=1$, meaning a coregionalization
matrix $\boldB_1$ of rank one. Samples share the same length-scale and
have similar form. They have different variances, though. Each sample
may be considered as a scaled version of the latent function, as it
can be seen from equation \ref{eq:lmc} with $Q=1$ and $R_q=1$,
\begin{align*}
f_1(x) & = a_{1,1}^1u^1_1(x), \qquad  f_2(x) = a_{2,1}^1u^1_1(x),
\end{align*}
where we have used $x$ instead of $\boldx$ for the one-dimensional
input space.

Figure \ref{fig:ICM:Rank2} shows samples from an ICM of rank two. From
equation \ref{eq:lmc}, we have for $Q=1$ and $R_q=2$,
\begin{align*}
f_1(x) & = a_{1,1}^1u^1_1(x) + a_{1,1}^2u^2_1(x), \qquad  f_2(x) = a_{2,1}^1u^1_1(x) + a_{2,1}^2u^2_1(x),
\end{align*}
where $u^1_1(x)$ and $u^2_1(x)$ are sampled from the same Gaussian
process. Outputs are weighted sums of two different latent functions
that share the same covariance. In contrast to the ICM of rank one, we
see from figure \ref{fig:ICM:Rank2} that both outputs have different
forms, although they share the same length-scale.

\begin{figure}[ht!]
\centering
\includegraphics[width=0.8\textwidth]{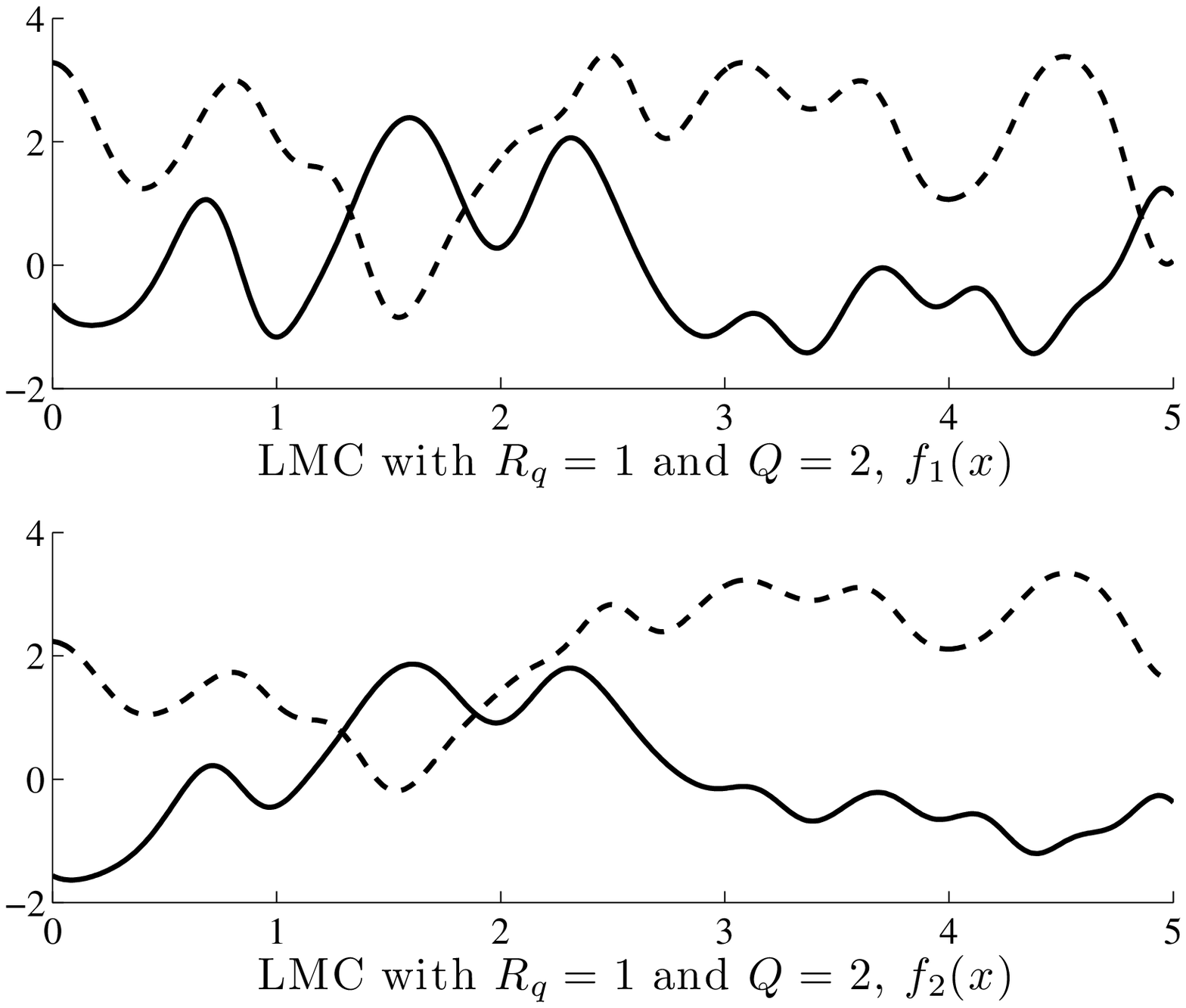}
\caption{Two samples from a linear model of coregionalization with
  $R_q=1$ and $Q=2$. The solid lines represent one of the samples. The dashed lines represent the other
  sample. Samples are the weigthed sums of latent functions with
  different length-scales.}
\label{fig:SLFM}
\end{figure}

Figure \ref{fig:SLFM} displays outputs sampled from a LMC with $R_q=1$
and two latent functions ($Q=2$) with different length-scales. Notice
that both samples are combinations of two terms, a long length-scale
term and a short length-scale term. According to equation
\ref{eq:lmc}, outputs are given as
\begin{align*}
f_1(x) & = a_{1,1}^1u^1_1(x) + a_{1,2}^1u^1_2(x), \qquad  f_2(x) = a_{2,1}^1u^1_1(x) + a_{2,2}^1u^1_2(x),
\end{align*}
where $u^1_1(x)$ and $u^1_2(x)$ are samples from two Gaussian
processes with different covariance functions. In a similar way to the
ICM of rank one (see figure \ref{fig:ICM:Rank1}), samples from both
outputs have the same form, this is, they are aligned.

\begin{figure}[ht!]
\centering
\includegraphics[width=0.8\textwidth]{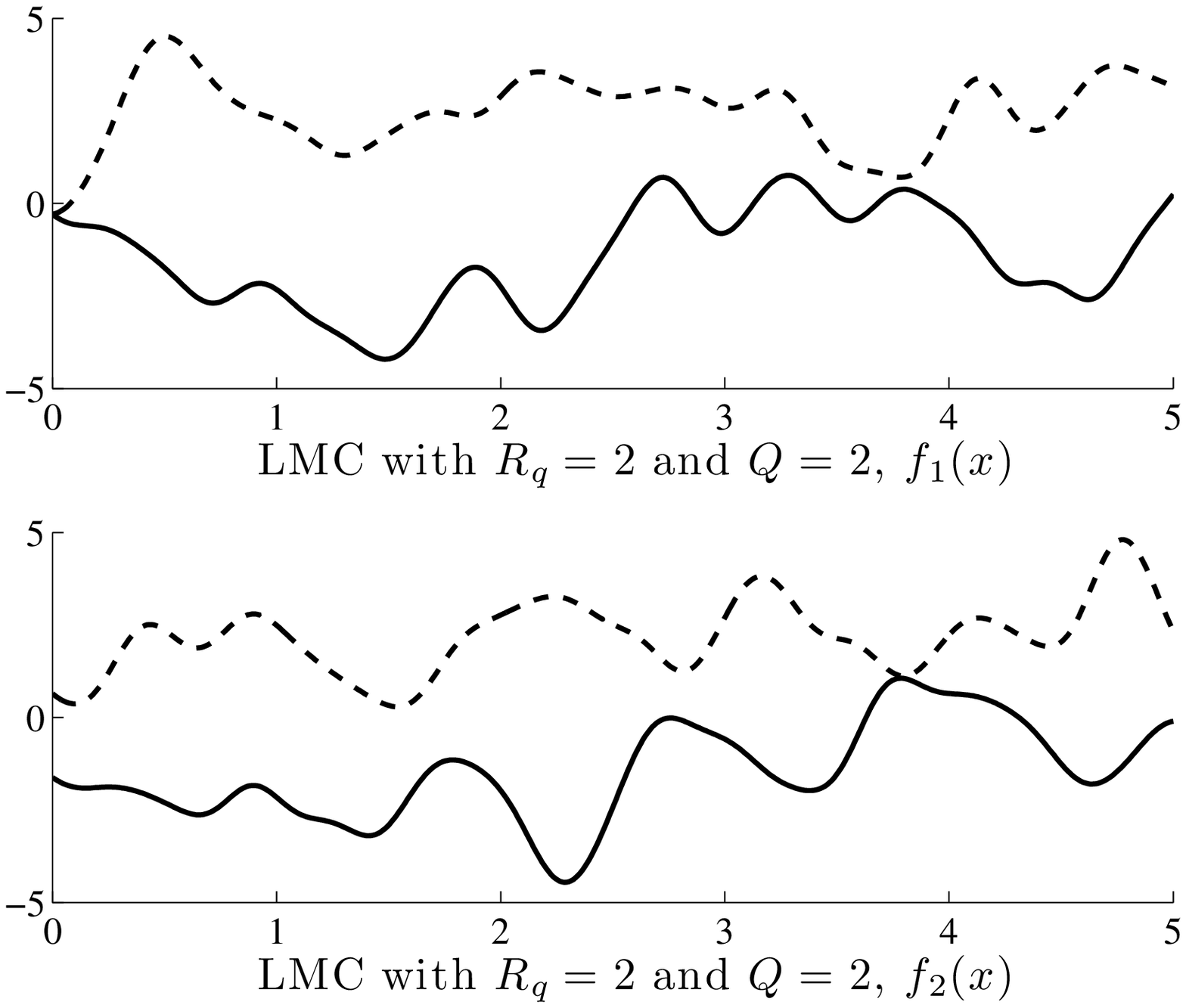}
\caption{Two samples from a linear model of coregionalization with
  $R_q=2$ and $Q=2$. The solid lines represent one of the samples. The dashed lines represent the other
  sample. Samples are the weigthed sums of four latent functions, two
  of them share a covariance with a long length-scale and the other
  two share a covariance with a shorter length-scale.}
\label{fig:LMC}
\end{figure}

We have the additional case for a LMC with $R_q=2$ and $Q=2$ in figure \ref{fig:LMC}. According to equation \ref{eq:lmc},
the outputs are give as
\begin{align*}
f_1(x) & = a_{1,1}^1u^1_1(x) + a_{1,1}^2u^2_1(x) + a_{1,2}^1u^1_2(x) + a_{1,2}^2u^2_2(x),\\
f_2(x) & = a_{2,1}^1u^1_1(x) + a_{2,1}^2u^2_1(x) + a_{2,2}^1u^1_2(x) + a_{2,2}^2u^2_2(x),
\end{align*}
where the pair of latent functions $u^1_1(x)$ and $u^2_1(x)$ share
their covariance function and the pair of latent functions $u^1_2(x)$
and $u^2_2(x)$ also share their covariance function. As in the case of
the LMC with $R_q=1$ and $Q=2$ in figure \ref{fig:SLFM}, the outputs
are combinations of a term with a long length-scale and a term with a
short length-scale. A key difference however, is that, for $R_q=2$ and
$Q=2$, samples from different outputs have different shapes.\footnote{Notice that samples from each output are not
synchronized, meaning that the maximums and minimus do not always occur at the same input points.}

\subsubsection{Linear Model of Coregionalization in Machine Learning and Statistics}\label{subsection:LMC:statistics:ML}

The linear model of coregionalization has already been used in
machine learning in the context of Gaussian processes for multivariate regression and in statistics for computer
emulation of expensive multivariate computer codes.

As we have seen before, the linear model of coregionalization imposes
the correlation of the outputs explicitly through the set of
coregionalization matrices. A simple idea used in the early papers of
multi-output GPs for machine learning was based on the intrinsic
coregionalization model and assumed $\boldB =\eye_D$. In other words,
the outputs were considered to be conditionally independent given the
parameters $\bm{\phi}$. Correlation between the outputs was assumed to
exist implicitly by imposing the same set of hyperparameters
$\bm{\phi}$ for all outputs and estimating those parameters, or
the kernel matrix $\Kms$ directly, using data from all the outputs
\cite{Minka:learnHowTo:1997,Lawrence:learningToLearn:2004,
  KaiYu:GPMT:2005}.

In this section, we review more recent approaches for multiple output modeling that are different versions of the linear
model of coregionalization.

\paragraph{Semiparametric latent factor model.} The semiparametric
latent factor model (SLFM) proposed by \cite{Teh:semiparametric05}
turns out to be a simplified version of the LMC. In fact it
corresponds to setting $R_q=1$ in \eqref{eq:lmc} so that we can
rewrite equation \eqref{eq:matricesLMC} as
\begin{align}\notag
\Kmv &= \sum_{q=1}^Q\bolda_{q}\bolda^{\top}_{q} \otimes k_q(\boldX, \boldX),
\end{align}
where $\bolda_{q}\in\rone^{D\times 1}$ with elements $\{a_{d,q}\}_{d=1}^D$ and $q$ fixed. With some algebraic manipulations,
that exploit the properties of the Kronecker product, we can write
\begin{align}\notag
\Kmv &= \sum_{q=1}^Q(\bolda_{q}\otimes \boldI_N)k_q(\boldX, \boldX)(\bolda^{\top}_{q} \otimes \boldI_N)
=(\boldAtilde \otimes \boldI_N)\mathbf{\widetilde{K}}(\boldAtilde^{\top} \otimes \boldI_N),
\end{align}
where $\boldAtilde\in\rone^{D\times Q}$ is a matrix
with columns $\bolda_q$ and $\mathbf{\widetilde{K}}\in\rone^{QN\times QN}$
is a block diagonal matrix with blocks given by $k_q(\boldX, \boldX)$.

The functions $u_q(\boldx)$ are considered to be latent factors and
the semiparametric name comes from the fact that it is combining a
nonparametric model, that is a Gaussian process, with a parametric
linear mixing of the functions $u_{q}(\boldx)$. The kernels $k_q$, for
each basic process is assumed to be \rbfKernelLong\ with a different
characteristic length-scale for each input dimension.  The informative
vector machine (IVM) \cite{Lawrence:ivm02} is employed to speed up
computations.

\paragraph{Gaussian processes for Multi-task,  Multi-output and Multi-class}
The intrinsic coregionalization model is considered by
\cite{Bonilla:multi07} in the context of multitask learning. The
authors use a probabilistic principal component analysis
(\textrm{PPCA}) model to represent the matrix $\boldB$.  The spectral
factorization in the \textrm{PPCA} model is replaced by an incomplete
Cholesky decomposition to keep numerical stability. The authors also
refer to the autokrigeability effect as the cancellation of inter-task
transfer \cite{Bonilla:multi07}, and discuss the similarities between
the multi-task GP and the ICM, and its relationship to the SLFM and
the LMC.

The intrinsic coregionalization model has also been used by
\cite{Rogers:towards08}. Here the matrix $\boldB$ is assumed to have a
spherical parametrization, $\boldB = \diag(\mathbf{e})\mathbf{S}
^{\top}\mathbf{S}\diag(\mathbf{e})$, where $\mathbf{e}$ gives a
description for the scale length of each output variable and
$\mathbf{S}$ is an upper triangular matrix whose $i$-th column is
associated with particular spherical coordinates of points in
$\rone^i$ (for details see sec. 3.4
\cite{osborne:multiGPsReport:2007}). The scalar kernel $k$ is
represented through a Matérn kernel, where different parameterizations
allow the expression of periodic and non-periodic terms.
Sparsification for this model is obtained using an IVM style approach.

In a classification context, Gaussian processes methodology has been
mostly restricted to the case where the outputs are conditionally
independent given the hyperparameters $\bm{\phi}$
\cite{Minka:learnHowTo:1997,Williams:multiclassGPs:1998,
  Lawrence:learningToLearn:2004,
  Seeger:sparseGpMultiClass:2004,KaiYu:GPMT:2005,
  Rasmussen:book06}. Therefore, the kernel matrix $\Kmv$ takes a
block-diagonal form, with blocks given by $\Kmsd{d}{d}$.  Correlation
between the outputs is assumed to exist implicitly by imposing the
same set of hyperparameters $\bm{\phi}$ for all outputs and estimating
those parameters, or directly the kernel matrices $\Kmsd{d}{d}$, using
data from all the outputs
\cite{Minka:learnHowTo:1997,Lawrence:learningToLearn:2004,
  KaiYu:GPMT:2005, Rasmussen:book06}. Alternatively, it is also
possible to have parameters $\bm{\phi}_d$ associated to each output
\cite{Williams:multiclassGPs:1998,Seeger:sparseGpMultiClass:2004}.

Only recently, the intrinsic coregionalization model has been used in
the multiclass scenario. In \cite{Skolidis:multiclassICM2011}, the
authors use the intrinsic coregionalization model for classification,
by introducing a probit noise model as the likelihood. Since the
posterior distribution is no longer analytically tractable, the
authors use Gibbs sampling, Expectation-Propagation (EP) and
variational Bayes \footnote{Mathematical treatment for each of these
  inference methods can be found in \cite{Bishop:PRLM06}, chapters 10
  and 11.} to approximate the distribution.

\paragraph{Computer emulation.} A computer emulator is a statistical
model used as a surrogate for a computationally expensive
deterministic model or computer code, also known as a
simulator. Gaussian processes have become the preferred statistical
model among computer emulation practitioners (for a review see
\cite{Ohagan:computerCodeTutorial:2006}).  Different Gaussian process
emulators have been recently proposed to deal with several outputs
\cite{Higdon:high08,Conti:multi09,Rougier:multi08,
  McFarland:multiEmulator:Single:2008,Bayarri:multiEmulator:single:2009,
  Quian:quan:qual:factor:2008}.

In \cite{Higdon:high08}, the linear model of coregionalization is used
to model images representing the evolution of the implosion of steel
cylinders after using TNT and obtained employing the so called
Neddemeyer simulation model (see \cite{Higdon:high08} for further
details).  The input variable $\boldx$ represents parameters of the
simulation model, while the output is an image of the radius of the
inner shell of the cylinder over a fixed grid of times and angles. In
the version of the LMC that the authors employed, $R_q=1$ and the $Q$
vectors $\bolda_q$ were obtained as the eigenvectors of a PCA
decomposition of the set of training images.

In \cite{Conti:multi09}, the intrinsic coregionalization model is
employed for emulating the response of a vegetation model called the
Sheffield Dynamic Global Vegetation Model (SDGVM)
\cite{Woodward:SDGVM:1998}. Authors refer to the ICM as the
Multiple-Output (MO) emulator. The inputs to the model are ten
($p=10$) variables related to broad soil, vegetation and climate data,
while the outputs are time series of the net biome productivity (NBP)
index measured at a particular site in a forest area of Harwood,
UK. The NBP index accounts for the residual amount of carbon at a
vegetation site after some natural processes have taken place. In the
paper, the authors assume that the outputs correspond to the different
sampling time points, so that $D=T$, being $T$ the number of time
points, while each observation corresponds to specific values of the
ten input variables. Values of the input variables are chosen
according to a maxi-min Latin hypercube design.

Rougier \cite{Rougier:multi08} introduces an emulator for multiple-outputs
that assumes that the set of output variables can be seen as a single
variable while augmenting the input space with an additional index
over the outputs. In other words, it considers the output variable as
an input variable.  \cite{Conti:multi09}, refers to the model in
\cite{Rougier:multi08} as the Time Input (TI) emulator and discussed
how the TI model turns out to be a particular case of the MO model
that assumes a particular \rbfKernelLong\ kernel (see chapter 4
\cite{Rasmussen:book06}) for the entries in the coregionalization
matrix $\boldB$.

McFarland \emph{et al.} \cite{McFarland:multiEmulator:Single:2008}
consider a multiple-output problem as a single output one. The setup
is similar to the one used in \cite{Conti:multi09}, where the number
of outputs are associated to different time points, this is, $D =
T$. The outputs correspond to the time evolutions of the temperature
of certain location of a container with decomposing foam, as function
of five different \emph{calibration} variables (input variables in
this context, $p=5$). The authors use the time index as an input (akin
to \cite{Rougier:multi08}) and apply a greedy-like algorithm to select
the training points for the Gaussian process. Greedy approximations
like this one have also been used in the machine learning literature
(for details, see \cite{Rasmussen:book06}, page 174).

Similar to \cite{Rougier:multi08} and
\cite{McFarland:multiEmulator:Single:2008}, Bayarri \emph{et al.}
\cite{Bayarri:multiEmulator:single:2009} use the time index as an
input for a computer emulator that evaluates the accuracy of CRASH, a
computer model that simulates the effect of a collision of a vehicle
with different types of barriers.

Quian \emph{et al.} \cite{Quian:quan:qual:factor:2008} propose a
computer emulator based on Gaussian processes that supports
quantitative and qualitative inputs. The covariance function in this
computer emulator is related to the ICM in the case of one qualitative
factor: the qualitative factor is considered to be the index of the
output, and the covariance function takes again the form
$k(\boldx,\boldx')k_T(d,d')$. In the case of more than one qualitative
input, the computer emulator could be considered a multiple output GP
in which each output index would correspond to a particular
combination of the possible values taken by the qualitative
factors. In this case, the matrix $\boldB$ in ICM would have a block
diagonal form, each block determining the covariance between the
values taken by a particular qualitative input.

\subsection{Extensions}\label{separable:kernels:extensions}
In this section we describe further developments related to the setting of separable kernels or SoS kernels, both from a
regularization and a Bayesian perspective.

\subsubsection{Extensions Within the Regularization Framework}
When we consider kernels of the form $\vk(\boldx, \boldx')=k(\boldx,
\boldx')\boldB$, a natural question is whether the matrix $\boldB$ can
be learned from data.  In a regression setting, one idea is to
estimate $\boldB$ in a separate inference step as the covariance
matrix of the output vectors in the training set and this is standard
in the geostatistics literature \cite{Wackernagel:book03}. A further question is whether we can learn both
$\boldB$ and an estimator within a unique inference step.  This is the
question tackled in \cite{jabave08}. The authors consider a variation
of the regularizer in \eqref{eq:clustered} and try to learn the
cluster matrix as a part of the optimization process.  More precisely
the authors considered a regularization term of the form
\beeq{eq:clustered2}{ R(\boldf) = \epsilon_1 \nor{\overline{f}}_k +
  \epsilon_2 \sum_{c=1}^r m_c \nor{\overline{f}_c-\overline{f}}^2_k+
  \epsilon_3 \sum_{c=1}^r \sum_{l \in I(c)} \nor{f^l -
    \overline{f}_c}^2_k,} where we recall that $r$ is the number of
clusters.  The three terms in the functional can be seen as: a global
penalty, a term penalizing {\em between cluster} variance and a term
penalizing {\em within cluster} variance. As in the case of the
regularizer in \eqref{eq:clustered}, the above
regularizer is completely characterized by a cluster matrix $\boldM$,
i.e. $R(\boldf)=R_\boldM(\boldf)$ (note that the corresponding matrix
$\boldB$ will be slightly different from \eqref{graphreg}).

 The idea is then to consider  a regularized functional
\begin{align}\label{tikhofuncvec}
\sum_{i=1}^D\frac{1}{N}\sum_{i=1}^N (f_j(\boldx_i)-y_{j,i})^2+\la R_\boldM(\boldf)
\end{align}
to be minimized jointly over $\boldf$ and $\boldM$ (see
\cite{jabave08} for details). This problem is typically non tractable
from a computational point of view, so the authors in \cite{jabave08}
propose a relaxation of the problem which can be shown to be convex.

A different approach is taken in \cite{arevpo08} and \cite{armapo08}.
In this case the idea is that only a a small subset of features is
useful to learn all the components/tasks.  In the simplest case the
authors propose to minimize a functional of the form
$$
 \sum_{d =1}^D  \left\{\frac{1}{N} \sum_{i=1}^N ( \boldw_d^\top U^\top\boldx_i-y_{d,i})^2+\la \boldw_d^\top\boldw_d\right\}.
$$
over $\boldw_1, \dots, \boldw_D \in \rone^{p}, U\in \rone^{D\times D}$
under the constraint $\text{Tr}(U_t^\top U_t)\le \gamma$.  Note that
the minimization over the matrix $U$ couples the otherwise disjoint
component-wise problems.  The authors of \cite{arevpo08} discuss how
the above model is equivalent to considering a kernel of the form
$$
K(\boldx, \boldx')=k_\boldD(\boldx, \boldx')\boldI_D, \quad\quad k_\boldD(\boldx, \boldx')=\boldx^\top \boldD\boldx'
$$
where $\boldD$ is a positive definite matrix and
a  model which  can be described components wise as
$$
f_d(\boldx)=\sum_{i=1}^pa_{d,i}x_j=\bolda_d^\top \boldx,
$$
making apparent the connection with the LMC model.
In fact, it is possible to show that
the above minimization problem is equivalent to minimizing
\begin{align}\label{tikho2}
\sum_{d=1}^D \frac{1}{N} \sum_{i=1}^N (\bolda_d^\top\boldx_i-y_{d,i})^2+\la \sum_{d=1}^D\bolda_d^\top \boldD\bolda_d,
\end{align}
over $\bolda'_1, \dots, \bolda'_D \in \rone^{p}$ and
$\text{Tr}(\boldD)\le 1$, where the last restriction is a convex
approximation of the low rank requirement.  Note that from a Bayesian
perspective the above scheme can be interpreted as learning a
covariance matrix for the response variables which is optimal for all
the tasks.  In \cite{arevpo08}, the authors consider a more general
setting where $\boldD$ is replaced by $F(\boldD)$ and show that if the
matrix valued function $F$ is matrix concave, then the induced
minimization problem is jointly convex in $(\bolda_i)$ and
$\boldD$. Moreover, the authors discuss how to extend the above
framework to the case of more general kernel functions. Note that an
approach similar to the one we just described is at the basis of
recent work exploiting the concept of sparsity while solving multiple
tasks. These latter methods cannot in general be cast in the framework
of kernel methods and we refer the interested reader to
\cite{obozinski2010joint} and references therein.

For the reasoning above the key assumption is that a response variable
is either important for all the tasks or not.  In practice it is
probably often the case that only certain subgroups of tasks share the
same variables.  This idea is at the basis of the study in
\cite{armapo08}, where the authors design an algorithm to learn at
once the group structure and the best set of variables for each groups
of tasks.  Let ${\cal G}=(G_t)_{t=1}^\top$ be a partition of the set
of components/tasks, where $G_t$ denotes a group of tasks and $|G_t|
\le D$.  Then the author propose to consider a functional of the form
\begin{align*}
\min_{\cal G}\sum_{G_t \in {\cal G}}
 \min_{\bolda_d, d \in G_t, U_t }
 \sum_{d \in G_t}\Bigg\{   \frac{1}{N}& \sum_{i=1}^N ( \bolda_d^\top U_t^\top\boldx_i-y_{d,i})^2
+\la \boldw_d^\top\boldw_d^\top\\&+\gamma \text{Tr}(U_t^\top U_t)\Bigg\},
\end{align*}
where $U_1, \dots U_T$ is a sequence of $p$ by $p$ matrices. The
authors show that while the above minimization problem is not convex,
stochastic gradient descent can be used to find local minimizers which
seems to perform well in practice.

\subsubsection{Extensions from the Gaussian Processes Perspective}

A recent extension of the linear model of coregionalization expresses
the output covariance function through a linear combination of
nonorthogonal latent functions \cite{Vargas:nonOrthogonal02}.  In
particular, the basic processes $u^i_q(\boldx)$ are assumed to be
nonorthogonal, leading to the following covariance function
\begin{align}\notag
  \cov[\boldf(\boldx),\boldf(\boldx')]&=\sum_{q=1}^Q\sum_{q'=1}^Q\boldB_{q,q'}k_{q,q'}(\boldx,\boldx'),
\end{align}
where $\boldB_{q,q'}$ are \emph{cross-coregionalization}
matrices. Cross-covariances $k_{q,q'}(\boldx, \boldx')$ can be
negative (while keeping positive semidefiniteness for
$\cov[\boldf(\boldx),\boldf(\boldx')]$), allowing negative
cross-covariances in the linear model of coregionalization.  The
authors argue that, in some real scenarios, a linear combination of
several correlated attributes are combined to represent a single model
and give examples in mining, hydrology and oil industry
\cite{Vargas:nonOrthogonal02}.

\section{Beyond Separable Kernels}\label{NonSepKer}
Working with separable kernels or SoS kernels is appealing for their simplicity, but
can be limiting in several applications. Next we review different
types of kernels that go beyond the separable case or SoS case.
\subsection{Invariant Kernels}
\paragraph{Divergence free and curl free fields.}
The following two kernels are matrix valued \rbfKernelLong\
(\rbfKernel) kernels \cite{narcowich1994generalized} and can be used
to estimate divergence-free or curl-free vector fields
\cite{macedo08divcurlfree} when the input and output space have the
same dimension.  These kernels induce a similarity between the vector
field components that depends on the input points, and therefore
cannot be reduced to the form $\vk(\boldx,\boldx') =
k(\boldx,\boldx')\boldB$.

We consider the case of vector fields with $D=p$, where ${\cal
  X}=\R^p$.  The divergence-free matrix-valued kernel can be defined
via a translation invariant matrix-valued \rbfKernel\ kernel
\[
\Phi(u) = (\nabla \nabla^\top - \nabla^\top \nabla I)\phi(u) = H\phi(u) - \operatorname{tr}(H\phi(u))\boldI_D \; ,
\]
where $H$ is the Hessian operator and $\phi$ a scalar \rbfKernel\
kernel, so that $\vk(\boldx, \boldx') := \Phi(\boldx-\boldx')$.

The columns of the matrix valued \rbfKernel\ kernel, $\Phi$, are
divergence-free. In fact, computing the divergence of a linear
combination of its columns, $\nabla^\top (\Phi(u)c)$, with $c \in
\rone^p$, it is possible to show that \cite{BRBV10}
$$
\nabla^\top (\Phi(u)c) =  (\nabla^\top \nabla \nabla^\top \phi(u)) c - (\nabla^\top \nabla^\top \nabla \phi(u)) c = 0 \; ,
$$
where the last equality follows applying the product rule of the
gradient, the fact that the coefficient vector $c$ does not depend
upon $u$ and the equality $a^\top aa^\top = a^\top a^\top a, \forall a
\in \rone^p$.

Choosing a \rbfKernelLong, we obtain the divergence-free kernel
\beeq{div}{
\vk(\boldx,\boldx') =
\frac{1}{\sigma^2}e^{-\frac{\nor{\boldx - \boldx'}^2}{2\sigma^2}}A_{\boldx,\boldx'},
}
where
\[
A_{\boldx,\boldx'}= \left(\negthickspace \left(
    \frac{\boldx-\boldx'}{\sigma}\right)\negthickspace
  \left(\frac{\boldx-\boldx'}{\sigma}\right)^\top \negthickspace
  +\left((p-1)-\frac{\nor{\boldx-\boldx'}^2}{\sigma^2}
  \right)\Id_p\negthickspace \right) \; .
\]
The curl-free matrix valued kernels are obtained as
\[
\vk(\boldx,\boldx'):= \Psi(\boldx-\boldx') =- \nabla \nabla^\top
\phi(\boldx-\boldx') = -H\phi(\boldx-\boldx') \; ,
\]
where $\phi$ is a scalar RBF.  It is easy to show that the columns of
$\Psi$ are curl-free. The $j$-th column of $\Psi$ is given by $\Psi
\bolde_j$, where $\bolde_j$ is the standard basis vector with a one in
the $j$-th position. This gives us
\[
\Phi_{cf}e_j = - \nabla \nabla^\top\Phi_{cf}e_j =
\nabla(-\nabla^\top\Phi_{cf}e_j) = \nabla g \; ,
\]
where $g = -\partial \phi / \partial x_j$. The function $g$ is a
scalar function and the curl of the gradient of a scalar function is
always zero.  Choosing a \rbfKernelLong, we obtain the following
curl-free kernel
\beeq{curl}{
\Gamma_{cf}(\boldx,\boldx') = \frac{1}{\sigma^2}e^{-\frac{\nor{\boldx - \boldx'}^2}{2\sigma^2}}\left(\Id_D - \left( \frac{\boldx-\boldx'}{\sigma}\right)\left(\frac{\boldx-\boldx'}{\sigma}\right)^\top\right) \; .
}
It is possible to consider a convex linear combination of these two
kernels to obtain a kernel for learning any kind of vector field,
while at the same time allowing reconstruction of the divergence-free
and curl-free parts separately (see \cite{macedo08divcurlfree}).  The
interested reader can refer to
\cite{narcowich1994generalized,lowitzsch2005density,fuselier2006refined}
for further details on matrix-valued RBF and the properties of
divergence-free and curl-free kernels.

\paragraph{Transformable kernels.}
Another  example of invariant kernels is discussed in \cite{capomi08} and
is given by kernels defined by transformations. For the purpose
of our discussion, let $\Y = \rone^D$, $\X_0$ be a Hausdorff space and $T_d$
a family of  maps (not necessarily linear) from $\X$ to $\X_0$
 for $d = \{1,\ldots,D\}$ .

Then, given a continuous scalar kernel $k: \X_0 \times \X_0 \to \rone$, it is possible to define the following matrix
 valued kernel for any $\boldx,\boldx' \in \X$
\[
\Big(\vk(\boldx,\boldx')\Big)_{d,d'} = k(T_d\boldx,T_{d'}\boldx').
\]
A specific instance of the above example is described by \cite{VW03} in
the context of system identification, see also  \cite{capomi08}  for further details.

\subsection{Further Extensions of the LMC}

In \cite{Gelfand:nonstationaryLMC04}, the authors introduced a
nonstationary version of the LMC, in which the coregionalization
matrices are allowed to vary as functions of the input variables. In
particular, $\boldB_q$ now depends on the input variable $\boldx$,
this is, $\boldB_q(\boldx, \boldx')$.  The authors refer to this model
as the spatially varying LMC (SVLMC). As a model for the varying
coregionalization matrix $\boldB_q(\boldx,\boldx')$, the authors
employ two alternatives. In one of them, they assume that
$\boldB_q(\boldx,\boldx')=(\bm{\alpha}(\boldx,\boldx'))^{\psi}\boldB_q$,
where $\bm{\alpha}(\boldx)$ is a covariate of the input variable, and
$\psi$ is a variable that follows a uniform prior. In the other
alternative, $\boldB_q(\boldx,\boldx')$ follows a Wishart spatial
process, which is constructed using the definition of a Wishart
distribution, as follows. Suppose $\boldZ\in \rone^{D\times P}$ is a
random matrix with entries $z_{d,p}\sim \gauss(0,1)$, independently
and identically distributed, for $d=1, \dots, D$ and $p=1, \ldots,
P$. Define the matrix $\bm{\Upsilon}=\bm{\Gamma}\boldZ$, with
$\bm{\Gamma}\in\rone^{D\times D}$. The matrix
$\bm{\Omega}=\bm{\Upsilon}\bm{\Upsilon}^{\top} =
\bm{\Gamma}\boldZ\boldZ^{\top} \bm{\Gamma}^{\top}\in \rone^{D\times
  D}$ follows a Wishart distribution
$\mathcal{W}(P,\bm{\Gamma}\bm{\Gamma}^{\top})$, where $P$ is known as
the \emph{number of degrees of freedom} of the distribution. The
spatial Wishart process is constructed assuming that $z_{d,p}$ depends
on the input $\boldx$, this is, $z_{d, p}(\boldx, \boldx')$, with
$z_{d, p}(\boldx, \boldx')\sim \gauss(0, \rho_d(\boldx, \boldx'))$,
where $\{\rho_d(\boldx, \boldx')\}_{d=1}^D$ are correlation functions.
Matrix $\bm{\Upsilon}(\boldx, \boldx')=\bm{\Gamma}\boldZ(\boldx,
\boldx')$ and $\bm{\Omega}(\boldx, \boldx')=\bm{\Upsilon}(\boldx,
\boldx')\bm{\Upsilon}^{\top}(\boldx, \boldx') =
\bm{\Gamma}\boldZ(\boldx, \boldx')\boldZ^{\top}(\boldx, \boldx')
\bm{\Gamma}^{\top}\in \rone^{D\times D}$ follows a spatial Wishart
process $\mathcal{SW}(P,\bm{\Gamma}\bm{\Gamma}^{\top},\{\rho_d(\boldx,
\boldx')\}_{d=1}^D )$. In \cite{Gelfand:nonstationaryLMC04}, authors
assume $\bm{\Gamma}=\boldI_D$ and $\rho_d(\boldx, \boldx')$ is the
same for all values of $d$. Inference in this model is accomplished
using Markov chain Monte Carlo. For details about the particular
inference procedure, the reader is referred to
\cite{Gelfand:nonstationaryLMC04}.

\subsection{Process Convolutions}\label{section:convolution}

More general non-separable kernels can also be constructed from a
generative point of view. We saw in section \ref{section:lmc} that the
linear model of coregionalization involves instantaneous mixing
through a linear weighted sum of independent processes to construct
correlated processes. By instantaneous mixing we mean that the output
function $\boldf(\boldx)$ evaluated at the input point $\boldx$ only
depends on the values of the latent functions
$\{u_q(\boldx)\}_{q=1}^Q$ at the same input $\boldx$.  This
instantaneous mixing leads to a kernel function for vector-valued
functions that has a separable form.

A non-trivial way to mix the latent functions is through convolving a
base process with a smoothing kernel.\footnote{We use kernel to refer
  to both reproducing kernels and smoothing kernels. Smoothing kernels
  are functions which are convolved with a signal to create a smoothed
  version of that signal.} If the base process is a Gaussian process,
it turns out that the convolved process is also a Gaussian process. We
can therefore exploit convolutions to construct covariance functions
\cite{Barry:balckbox96,verHoef:convolution98,Higdon:ocean98,Higdon:convolutions02,Boyle:dependent04,Lawrence:gpsim2007a,Alvarez:sparse2009}.

In a similar way to the linear model of coregionalization, we consider
$Q$ groups of functions, where a particular group $q$ has elements
$u^i_q(\mathbf{z})$, for $i=1,\ldots, R_q$. Each member of the group
has the same covariance $k_q(\boldx,\boldx')$, but is sampled
independently. Any output $f_d(\boldx)$ is described by
\begin{align*}
f_d(\mathbf{x})&=\sum_{q=1}^Q\sum_{i=1}^{R_q}\int_{\inputSpace}G^{i}_{d,q}(\mathbf{x}-\mathbf{z})u^i_q(\mathbf{z})\dif\mathbf{z}
+w_d(\boldx)=\sum_{q=1}^Qf^q_d(\boldx)+w_d(\boldx),
\end{align*}
where
\begin{align}
  f^q_d(\mathbf{x})&=\sum_{i=1}^{R_q}\int_{\inputSpace}G^{i}_{d,q}(\mathbf{x}-\mathbf{z})u^i_q(\mathbf{z})\dif\mathbf{z},
\label{eq:process:conv:plus:lmc:each:q}
\end{align}
and $\{w_d(\boldx)\}_{d=1}^D$ are independent Gaussian processes with
zero mean and covariance $k_{w_d}(\boldx, \boldx')$.  For the
integrals in equation \eqref{eq:process:conv:plus:lmc:each:q} to
exist, it is assumed that each kernel $G^i_{d,q}(\mathbf{x})$ is a
continuous function with compact support
\cite{Hormander:analysis:PDE:1983} or square-integrable
\cite{verHoef:convolution98,Higdon:convolutions02}. The kernel
$G^i_{d,q}(\mathbf{x})$ is also known as the moving average function
\cite{verHoef:convolution98} or the smoothing kernel
\cite{Higdon:convolutions02}. We have included the superscript $q$ for
$f^q_d(\mathbf{x})$ in \eqref{eq:process:conv:plus:lmc:each:q} to
emphasize the fact that the function depends on the set of latent
processes $\{u^i_q(\boldx)\}_{i=1}^{R_q}$.  The latent functions
$u^i_q(\mathbf{z})$ are Gaussian processes with general covariances
$k_q(\boldx,\boldx')$.

Under the same independence assumptions used in the linear model of
coregionalization, the covariance between $f_d(\boldx)$ and
$f_{d'}(\boldx')$ follows
\begin{align}
(\boldK(\boldx, \boldx'))_{d,d'}&=\sum_{q=1}^Qk_{f^q_d,f^q_{d'}}(\boldx, \boldx')+k_{w_d}(\boldx, \boldx')\delta_{d,d'},
\label{eq:covf}
\end{align}
where
\begin{align}
  k_{f^q_d,f^q_{d'}}(\boldx,
  \boldx')&=\sum_{i=1}^{R_q}\int_{\inputSpace}
  G^i_{d,q}(\mathbf{x}-\mathbf{z})\int_{\inputSpace}G^i_{d',q}(\mathbf{x'}-\mathbf{z'})k_q(\mathbf{z},\mathbf{z}')
  \dif\mathbf{z'}\dif\mathbf{z}.\label{eq:covfq}
\end{align}
Specifying $G^i_{d,q}(\boldx-\mathbf{z})$ and
$k_q(\mathbf{z},\mathbf{z}')$ in the equation above, the covariance
for the outputs $f_d(\mathbf{x})$ can be constructed
indirectly. Notice that if the smoothing kernels are taken to be the
Dirac delta function in equation \eqref{eq:covfq}, such that
$G^i_{d,q}(\mathbf{x}-\mathbf{z}) =
a^i_{d,q}\delta(\mathbf{x}-\mathbf{z})$,\footnote{We have slightly
  abused of the delta notation to indicate the Kronecker delta for
  discrete arguments and the Dirac function for continuous
  arguments. The particular meaning should be understood from the
  context.}  the double integral is easily solved and the linear model
of coregionalization is recovered.  In this respect, process
convolutions could also be seen as a dynamic version of the linear
model of coregionalization in the sense that the latent functions are
dynamically transformed with the help of the kernel smoothing
functions, as opposed to a static mapping of the latent functions in
the LMC case. See section \ref{section:comparison:pc:lmc} for a
comparison between the process convolution and the LMC.

A recent review of several extensions of this approach for the single
output case is presented in \cite{Calder:convolution07}.  Some of
those extensions include the construction of nonstationary covariances
\cite{Higdon:ocean98,Higdon:nonstationaryCov:1998,
  Fuentes:nonstationaryAirPollution:2002,
  Fuentes:spectralNonstationary:2002,Pacioreck:nonstationaryCov:2004}
and spatiotemporal covariances
\cite{Wikle:hierarchicalBayesSpaceTimeModels98,Wikle:kernelBasedSpectralModel02,
  Wikle:hierarchicalEcologicalProcesses03}.

The idea of using convolutions for constructing multiple output
covariances was originally proposed by
\cite{verHoef:convolution98}. They assumed that $Q=1$, $R_q=1$, that
the process $u(\boldx)$ was white Gaussian noise and that the input
space was $\inputSpace=\rone^p$. \cite{Higdon:convolutions02} depicted
a similar construction to the one introduced by
\cite{verHoef:convolution98}, but partitioned the input space into
disjoint subsets $\mathcal{X}=\bigcap_{d=0}^D\mathcal{X}_d$, allowing
dependence between the outputs only in certain subsets of the input
space where the latent process was common to all
convolutions.\footnote{The latent process $u(\boldx)$ was assumed to
  be independent on these separate subspaces.}

Higdon \cite{Higdon:convolutions02} coined the general moving average
construction to develop a covariance function in equation
\eqref{eq:process:conv:plus:lmc:each:q} as a \emph{process
  convolution}.

Boyle and Frean \cite{Boyle:dependent04,Boyle:techreport:2005}
introduced the process convolution approach for multiple outputs to
the machine learning community with the name of ``dependent Gaussian
processes'' (DGP), further developed in
\cite{Boyle:phdThesis:2007}. They allow the number of latent functions
to be greater than one ($Q\ge 1$). In \cite{Lawrence:gpsim2007a} and
\cite{Alvarez:sparse2009}, the latent processes
$\{u_q(\boldx)\}_{q=1}^Q$ followed a more general Gaussian process
that goes beyond the white noise assumption.

\subsubsection{Comparison Between Process Convolutions and LMC}\label{section:comparison:pc:lmc}

\begin{figure}
\centering
\includegraphics[width=0.99\textwidth]{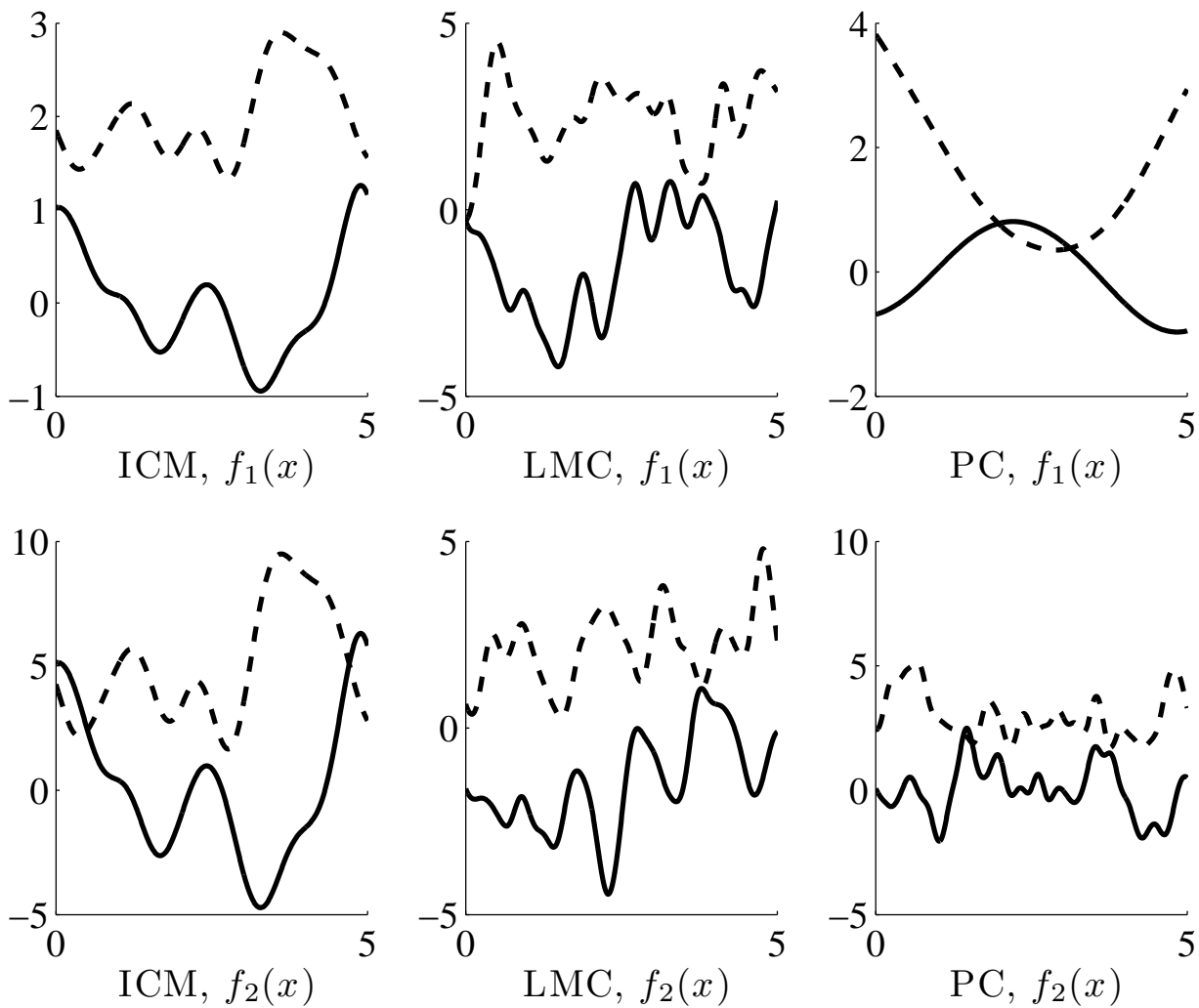}
\caption{Two samples from three kernel matrices obtained using the
  intrinsic coregionalization model with $R_q=1$ (first column), the
  linear model of coregionalization with $R_q=2$ (second column) and
  the process convolution formalism with $R_q=1$ and $Q=1$ (third
  column). Solid lines represent one of the samples. Dashed lines represent the other sample. There are
  two outputs, one row per output. Notice that for the ICM and the
  LMC, the outputs have the same length-scale (in the ICM case) or
  combined length-scales (in the LMC case).  The outputs generated
  from the process convolution covariance differ in their relative
  length-scale.}
\label{fig:toy:ICM:LMC:PC}
\end{figure}

Figure \ref{fig:toy:ICM:LMC:PC} shows an example of the instantaneous
mixing effect obtained in the ICM and the LMC, and the
non-instantaneous mixing effect due to the process convolution
framework. We sampled twice from a two-output Gaussian process with an
ICM covariance with $R_q=1$ (first column), an LMC covariance with
$R_q=2$ (second column) and a process convolution covariance with
$R_q=1$ and $Q=1$ (third column). As in the examples for the LMC, we
use \rbfKernel\ kernels for the basic kernels $k_q(\boldx,
\boldx')$. We also use an \rbfKernelLong form for the smoothing kernel
functions $G_{1,1}^1(\boldx-\boldx')$ and $G_{2,1}^1(\boldx-\boldx')$
and assume that the latent function is white Gaussian noise.

Notice from Figure \ref{fig:toy:ICM:LMC:PC} that samples from the ICM
share the same length-scale. Samples from the LMC are weighted sums of
functions with different length-scales (a long length-scale term and a
short length-scale term). In both models, ICM and LMC, the two outputs
share the same length-scale or the same combination of
length-scales. Samples from the PC show that the contribution from the
latent function is different over each output. Output $f_1(x)$ has a
long length-scale behavior, while output $f_2(x)$ has a short
length-scale behavior.

\begin{figure}[ht!]
\centering
\includegraphics[width=0.8\textwidth]{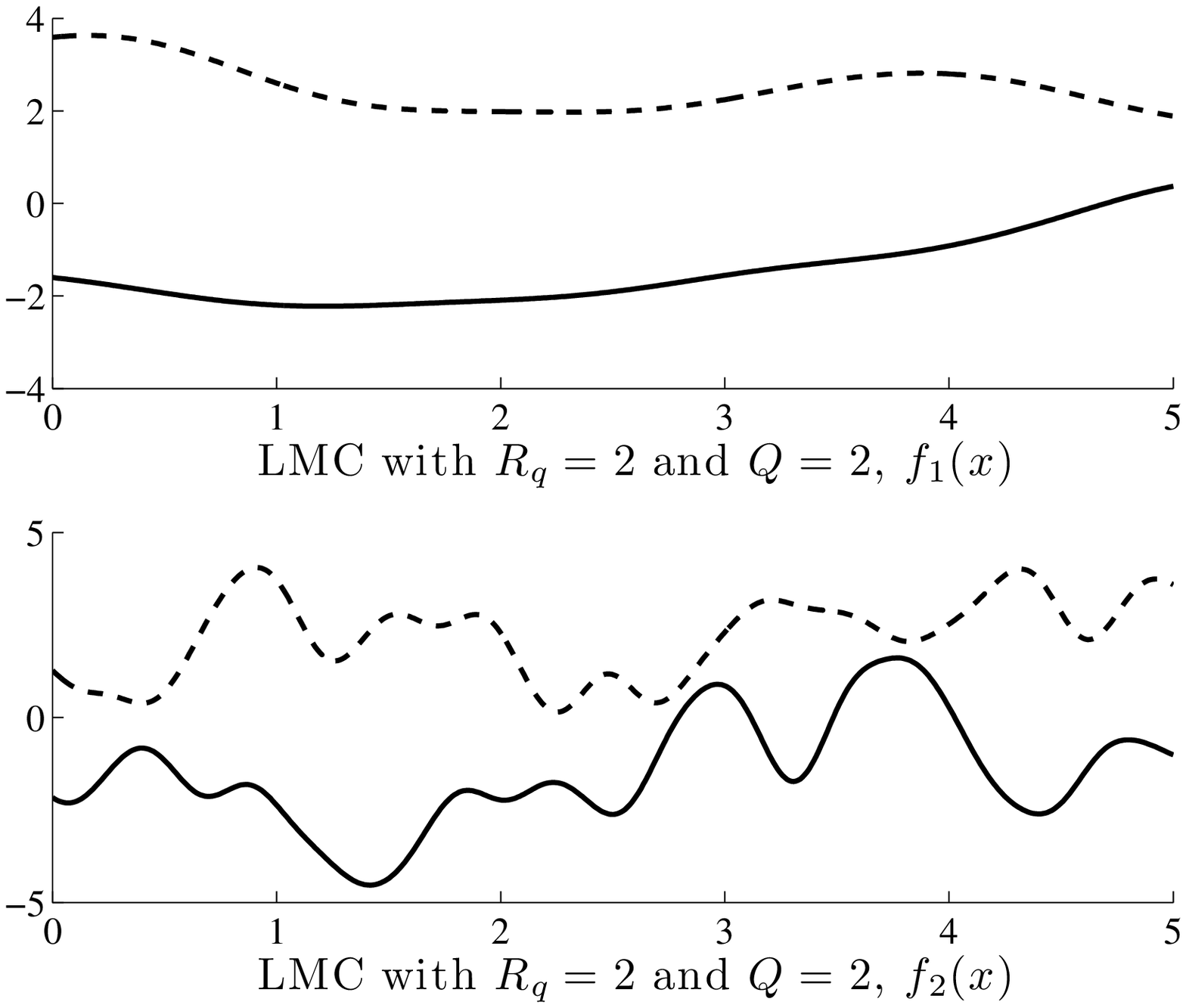}
\caption{Two samples from a linear model of coregionalization with
  $R_q=2$ and $Q=2$. The solid lines represent one of the samples. The dashed lines represent the other
  sample. Samples share a long length-scale behavior. An added short
  length-scale term appears only in output two.}
\label{fig:LMCConv}
\end{figure}

It would be possible to get similar samples to the PC ones using a
LMC. We would need to assume, though, that some covariances in a
particular coregionalization matrix $\boldB_q$ are zero. In Figure
\ref{fig:LMCConv}, we display samples from a LMC with $R_q=2$ and
$Q=2$. We have forced $b^2_{1,1}=b^2_{1,2}=b^2_{2,1}=0$. To generate
these samples we use an ICM with $R_q=2$ and a latent function with
long length-scale, and then add a sample from an independent Gaussian
process with a short length-scale to output $f_2(x)$.  It is debatable
if this compound model (ICM plus independent GP) would capture the
relevant correlations between the output functions.

To summarize, the choice of a kernel corresponds to specifying
dependencies among inputs and outputs.  In the linear model of
co-regionalization this is done considering separately inputs, via the
kernels $k_q$, and outputs, via the coregionalization matrices
$\boldB_q$, for $q=1,\dots, Q$.  Having a large large value of $Q$
allows for a larger expressive power of the model. For example if the
output components are substantially different functions (different
smoothness or length scale), we might be able to capture their
variability by choosing a sufficiently large $Q$. This is at the
expense of a larger computational burden.

On the other hand, the process convolution framework attempts to model
the variance of the set of outputs by the direct association of a
different smoothing kernel $G_d(\boldx)$ to each output $f_d(\boldx)$.
By specifying $G_d(\boldx)$, one can model, for example, the degree of
smoothness and the length-scale that characterizes each output. If
each output happens to have more than one degree of variation
(marginally, it is a sum of functions of varied smoothness) one is
faced with the same situation as in LMC, namely, the need to augment
the parameter space so as to satisfy a required precision. However,
due to the local description of each output that the process
convolution performs, it is likely that the parameter space for the
process convolution approach grows slower than the parameter space for
LMC.

\subsubsection{Other Approaches Related to Process Convolutions}

In \cite{Majumdar:convolvedCovariance07}, a different moving average
construction for the covariance of multiple outputs was introduced. It
is obtained as a convolution over covariance functions in contrast to
the process convolution approach where the convolution is performed
over processes. Assuming that the covariances involved are isotropic
and the only latent function $u(\boldx)$ is a white Gaussian noise,
\cite{Majumdar:convolvedCovariance07} show that the cross-covariance
obtained from
\begin{align*}
\cov\left[f_d(\mathbf{x}+\boldh),f_{d'}(\mathbf{x})\right] & =
\int_{\inputSpace}k_d(\mathbf{h}-\mathbf{z})k_{d'}(\mathbf{z})\dif\mathbf{z},
\end{align*}
where $k_d(\boldh)$ and $k_{d'}(\boldh)$ are covariances associated to
the outputs $d$ and $d'$, lead to a valid covariance function for the
outputs $\{f_d(\boldx)\}_{d=1}^D$. If we assume that the smoothing
kernels are not only square integrable, but also positive definite
functions, then the covariance convolution approach turns out to be a
particular case of the process convolution approach
(square-integrability might be easier to satisfy than positive
definiteness).

\cite{RMurray:transformationGPs:2005} introduced the idea of
transforming a Gaussian process prior using a discretized process
convolution, $\boldf_d = \boldG_d\boldu$, where
$\boldG_d\in\rone^{N\times M}$ is a so called \emph{design matrix}
with elements $\{\boldG_d(\boldx_n, \boldz_m)\}_{n=1,m=1}^{N,M}$ and
$\boldu^{\top}=[u(\boldx_1), \ldots, u(\boldx_M)]$. Such a
transformation could be applied for the purposes of fusing the
information from multiple sensors, for solving inverse problems in
reconstruction of images or for reducing computational complexity
working with the filtered data in the transformed space
\cite{Shi:LargeData:FilteredGPs:2005}.  Convolutions with general
Gaussian processes for modelling single outputs, were also proposed by
\cite{Fuentes:nonstationaryAirPollution:2002,
  Fuentes:spectralNonstationary:2002}, but instead of the continuous
convolution, \cite{Fuentes:nonstationaryAirPollution:2002,
  Fuentes:spectralNonstationary:2002} used a discrete convolution. The
purpose in \cite{Fuentes:nonstationaryAirPollution:2002,
  Fuentes:spectralNonstationary:2002} was to develop a spatially
varying covariance for single outputs, by allowing the parameters of
the covariance of a base process to change as a function of the input
domain.

Process convolutions are closely related to the Bayesian kernel method
\cite{Pillai:kernelHilbert07,Liang:bayesianKernelMethods09} construct
reproducible kernel Hilbert spaces (RKHS) by assigning priors to
signed measures and mapping these measures through integral
operators. In particular, define the following space of functions,
\begin{align}\notag
\mathcal{F}&=\Big\{f\Big|f(x)=\int_{\inputSpace}G(x,z)\gamma(\dif z),\;\gamma\in\Gamma\Big\},
\end{align}
for some space $\Gamma\subseteq\mathcal{B}(\inputSpace)$ of signed
Borel measures. In \cite[proposition~1] {Pillai:kernelHilbert07}, the
authors show that for $\Gamma=\mathcal{B}(\inputSpace)$, the space of
all signed Borel measures, $\mathcal{F}$ corresponds to a
RKHS. Examples of these measures that appear in the form of stochastic
processes include Gaussian processes, Dirichlet processes and Lévy
processes.  In principle, we can extend this framework for the
multiple output case, expressing each output as
\[
f_d(x)=\int_{\inputSpace}G_d(x,z)\gamma(\dif z).
\]
\section{Inference and Computational Considerations}\label{parEstimation}

Practical use of multiple-output kernel functions require the tuning
of the hyperparameters, and dealing with the issue of computational
burden related directly with the inversion of matrices of dimension
$ND\times ND$.  Cross-validation and maximization of the log-marginal
likelihood are alternatives for parameter tuning, while matrix
diagonalization and reduced-rank approximations are choices for
overcoming computational complexity of the matrix inversion.

In this section we refer to the parameter estimation problem for the
models presented in section \ref{SepKer} and \ref{NonSepKer} and also
to the computational complexity when using those models in practice.

\subsection{Estimation of Parameters in Regularization Theory}\label{parEstimation:reg}

From a regularization perspective, once the kernel is fixed, to find a
solution we need to solve the linear system defined in
\eqref{vectortikho}. The regularization parameter as well as the
possible kernel parameters are typically tuned via
cross-validation. The kernel free-parameters are usually reduced to
one or two scalars (e.g. the width of a scalar kernel). While
considering for example separable kernels the matrix $\boldB$ is fixed
by design, rather than learned, and the only free parameters are those
of the scalar kernel.

Solving problem \eqref{vectortikho}, this is $\boldcv=(\Kmv+\la N\eye
)^{-1}\boldyv$, is in general a costly operation both in terms of
memory and time.  When we have to solve the problem for a single value
of $\la$ Cholesky decomposition is the method of choice, while when we
want to compute the solution for different values of $\la$ (for
example to perform cross validation) singular valued decomposition
(SVD) is the method of choice.  In both case the complexity in the
worst case is $O(D^3N^3)$ (with a larger constant for the SVD) and the
associated storage requirement is $O(D^2N^2)$

 As observed in \cite{BRBV10}, this computational burden can be
 greatly reduced for separable kernels.  For example, if we consider
 the kernel $\vk(\boldx,\boldx')=k(\boldx,\boldx')\boldI$ the kernel
 matrix $\Kmv$ becomes block diagonal.  In particular if the input
 points are the same, all the blocks are equal and the problem reduces
 to inverting an $N$ by $N$ matrix.  The simple example above serves
 as a prototype for the more general case of a kernel of the form
 $\vk(\boldx,\boldx')=k(\boldx,\boldx')\boldB$.  The point is that for
 this class of kernels, we can use the eigen-system of the matrix
 $\boldB$ to define a new coordinate system where the kernel matrix
 becomes block diagonal.

 We start observing that if we denote with $(\sigma_1, \boldu_1),
 \dots, (\sigma_D, \boldu_D)$ the eigenvalues and eigenvectors of
 $\boldB$ we can write the matrix $\boldC=(\boldc_1,\dots, \boldc_N)$,
 with $\boldc_i \in \rone^D$, as $\boldC = \sum_{d=1}^D
 \tilde{\boldc}^d \otimes \boldu_d,$ where
 $\tilde{\boldc}^d=(\scal{\boldc_1}{\boldu_d}_D, \dots,
 \scal{\boldc_N}{\boldu_d}_D)$ and $\otimes$ is the tensor product and
 similarly $\tilde{\SY} = \sum_{d=1}^D \tilde{\boldy}^d \otimes
 \boldu_d,$ with $\tilde{\boldy}^d=(\scal{\boldy_1}{\boldu_d}_D,
 \dots, \scal{\boldy_N}{\boldu_d}_D)$.  The above transformations are
 simply rotations in the output space.  Moreover, for the considered
 class of kernels, the kernel matrix $\Kmv$ is given by the tensor
 product of the $N\times N$ scalar kernel matrix $\Kms$ and $\boldB$,
 that is $\Kmv = \boldB \otimes\Kms $.

Then  we have the following equalities
\begin{align*}
  \vC = (\Kmv+N\lambda_N\boldI)^{-1}\vY &=
  \sum_{d=1}^D (\boldB \otimes \Kms +N\lambda_N \boldI)^{-1} \tilde{\boldy}^d \otimes \boldu_d\\
  &= \sum_{d=1}^D (\sigma_d \Kms+N\lambda_N \boldI )^{-1}
  \tilde{\boldy}^d \otimes \boldu_d.
\end{align*}

Since the eigenvectors $\boldu_j$ are orthonormal, it follows that:
\beeq{trick}{ \tilde{\boldc}^d = (\sigma_d \Kms+N\lambda_N \boldI
  )^{-1} \tilde{\boldy}^j=\left( \Kms +\frac{\la_N}{\sigma_d} \boldI\right)^{-1}
  \frac{\tilde{\boldy}^d}{\sigma_d}, } for $d = 1, \ldots, D$. The
above equation shows that in the new coordinate system we have to
solve $D$ essentially independent problems after rescaling each kernel
matrix by $\sigma_d$ or equivalently rescaling the regularization
parameter (and the outputs).  The above calculation shows that all
kernels of this form allow for a simple implementation at the price of
the eigen-decomposition of the matrix $\boldB$.  Then we see that the
computational cost is now essentially $O(D^3)+O(N^3)$ as opposed to
$O(D^3N^3)$ in the general case.  Also, it shows that the coupling
among the different tasks can be seen as a rotation and rescaling of
the output points. Stegle \emph{et al.} \cite{Stegle:sparse11} also applied
this approach in the context of fitting matrix variate Gaussian models
with spherical noise.

\subsection{Parameters Estimation for Gaussian Processes}

In machine learning parameter estimation for Gaussian processes is
often approached through maximization of the marginal likelihood. The
method also goes by the names of evidence approximation, type II
maximum likelihood, empirical Bayes, among others
\cite{Bishop:PRLM06}.

With a Gaussian likelihood and after integrating $\boldf$ using the
Gaussian prior, the marginal likelihood is given by
\begin{align}\label{eq:marginal:likelihood}
  p(\mathbf{y}|\mathbf{X},\bm{\phi}) =
  \mathcal{N}(\boldy|\mathbf{0},\Kmv+\bm{\Sigma}),
\end{align}
where $\bm{\phi}$ are the hyperparameters.

The objective function is the logarithm of the marginal likelihood
\begin{align}
  \log p(\mathbf{y}|\mathbf{X},\bm{\phi})=
  -\frac{1}{2}\boldy^\top(\Kmv+&\bm{\Sigma})^{-1}\boldy
  -\frac{1}{2}\log|\Kmv+\bm{\Sigma}|\notag\\
  &-\frac{ND}{2}\log 2\pi.\label{eq:log:marginal:likelihood}
\end{align}
The parameters $\bm{\phi}$ are obtained by maximizing $\log
p(\mathbf{y}|\mathbf{X},\bm{\phi})$ with respect to each element in
$\bm{\phi}$. Maximization is performed using a numerical optimization
algorithm, for example, a gradient based method.  Derivatives follow
\begin{align}\notag
  \frac{\partial \log p(\mathbf{y}|\mathbf{X},\bm{\phi})}{\partial
    \phi_i}&=\frac{1}{2}\boldy^\top \overline{\Kmv}^{-1}\frac{\partial
    \overline{\Kmv}}{\partial \phi_i}\overline{\Kmv}^{-1}\boldy
  \\
  &- \frac{1}{2}\tr\left(\overline{\Kmv}^{-1}\frac{\partial
      \overline{\Kmv}}{\partial
      \phi_i}\right),\label{eq:marginal:likelihood:gradients}
\end{align}
where $\phi_i$ is an element of the vector $\bm{\phi}$ and
$\overline{\boldK(\boldX, \boldX)}=\boldK(\boldX,
\boldX)+\bm{\Sigma}$.  In the case of the LMC, in which the
coregionalization matrices must be positive semidefinite, it is
possible to use an incomplete Cholesky decomposition
$\boldB_q=\widetilde{\mathbf{L}}_q\widetilde{\mathbf{L}}_q^{\top}$,
with $\widetilde{\mathbf{L}}_q\in\rone^{D\times R_q}$, as suggested in
\cite{Bonilla:multi07}. The elements of the matrices $\mathbf{L}_q$
are considered part of the vector $\bm{\phi}$.

Another method used for parameter estimation, more common in the
geostatistics literature, consists of optimizing an objective function
which involves some empirical measure of the correlation between the
functions $f_d(\boldx)$, $\widehat{\boldK}(\boldx,\boldx')$, and the
multivariate covariance obtained using a particular model,
$\boldK(\boldx,\boldx')$ \cite{Goulard:FittingLMC92,
  Kunsch:generalizedCrossCovariance97,Pelletier:GLSLMC04}. Assuming
stationary covariances, this criteria reduces to
\begin{align}\label{eq:cost:function:LMC}
\mbox{WSS} = \sum_{i=1}^Nw(\boldh_i)\tr\left\{\left[\left(\widehat{\boldK}(\boldh_i)-
\boldK(\boldh_i)\right)\right]^{2}\right\},
\end{align}
where $\boldh_i = \boldx_i-\boldx'_i$ is a lag vector, $w(\boldh_i)$
is a weight coefficient, $\widehat{\boldK}(\boldh_i)$ is an
experimental covariance matrix with entries obtained by different
estimators for cross-covariance functions
\cite{Papritz:pseudoCrosvariogram93,verHoef:convolution98}, and
$\boldK(\boldh_i)$ is the covariance matrix obtained, for example,
using the linear model of coregionalization.\footnote{Note that the
  common practice in geostatistics is to work with variograms instead
  of covariances. A variogram characterizes a general class of random
  functions known as intrinsic random functions
  \cite{Matheron:variogramDef73}, which are random processes whose
  increments follow a stationary second-order process. For clarity of
  exposition, we will avoid the introduction of the variogram and its
  properties. The interested reader can follow the original paper by
  \cite{Matheron:variogramDef73} for a motivation of their existence,
  \cite{Gneiting:variogramCovariance01} for a comparison between
  variograms and covariance functions and \cite{Goovaerts:book97} for
  a definition of the linear model of coregionalization in terms of
  variograms.} One of the first algorithms for estimating the
parameter vector $\bm{\phi}$ in LMC was proposed by
\cite{Goulard:FittingLMC92}. It assumed that the parameters of the
basic covariance functions $k_q(\boldx, \boldx')$ had been determined
a priori and then used a weighted least squares method to fit the
coregionalization matrices.  In \cite{Pelletier:GLSLMC04} the
efficiency of other least squares procedures was evaluated
experimentally, including ordinary least squares and generalized least
squares. Other more general algorithms in which all the parameters are
estimated simultaneously include simulated annealing
\cite{Lark:simulatedAnnealingLMC03} and the EM algorithm
\cite{Zhang:EMLMC07}. Ver Hoef and Barry \cite{verHoef:convolution98} also proposed the
use of an objective function like \eqref{eq:cost:function:LMC}, to
estimate the parameters in the covariance obtained from a process
convolution.

Both methods described above, the evidence approximation or the least-square method, give point estimates of the
parameter vector $\bm{\phi}$. Several authors have employed full Bayesian inference by assigning priors to $\bm{\phi}$
and computing the posterior distribution through some sampling procedure. Examples include \cite{Higdon:high08} and
\cite{Conti:multi09} under the LMC framework or \cite{Boyle:dependent04} and \cite{Titsias:control:vars:2009} under
the process convolution approach.

As mentioned before, for non-Gaussian likelihoods, there is not a
closed form solution for the posterior distribution nor for the
marginal likelihood. However, the marginal likelihood can be
approximated under a Laplace, variational Bayes or expectation
propagation (EP) approximation frameworks for multiple output
classification \cite{Skolidis:multiclassICM2011,
  Bonilla:preferenceICM11}, and used to find estimates for the
hyperparameters. Hence, the error function is replaced for $\log
q(\mathbf{y}|\mathbf{X},\bm{\phi})$, where
$q(\mathbf{y}|\mathbf{X},\bm{\phi})$ is the approximated marginal
likelihood. Parameters are again estimated using a gradient based
methods.

The problem of computational complexity for Gaussian processes in the
multiple output context has been studied by different authors
\cite{Rougier:multi08, verHoelf:convolutionFFT04,
  Teh:semiparametric05, Boyle:phdThesis:2007, Alvarez:sparse2009,
  Alvarez:phdThesis:2011}. Fundamentally, the computational problem is
the same than the one appearing in regularization theory, that is, the
inversion of the matrix $\overline{\Kmv}=\Kmv+\bm{\Sigma}$ for solving
equation \eqref{vectortikho}. This step is necessary for computing the
marginal likelihood and its derivatives (for estimating the
hyperparameters as explained before) or for computing the predictive
distribution. With the exception of the method by
\cite{Rougier:multi08}, the approximation methods proposed in
\cite{verHoelf:convolutionFFT04, Teh:semiparametric05,
  Boyle:phdThesis:2007, Alvarez:sparse2009, Alvarez:phdThesis:2011}
can be applied to reduce computational complexity, whichever
covariance function (LMC or process convolution, for example) is used
to compute the multi-output covariance matrix. In other words, the
computational efficiency gained is independent of the particular
method employed to compute the covariance matrix.

Before looking with some detail at the different approximation methods
employed in the Gaussian processes literature for multiple outputs, it
is worth mentioning that computing the kernel function through process
convolutions in equation \eqref{eq:covfq} implies solving a double
integral, which is not always feasible for any choice of the smoothing
kernels $G_{d,q}^i(\cdot)$ and covariance functions
$k_q(\boldx,\boldx')$. An example of an analytically tractable
covariance function occurs when both the smoothing kernel and the
covariance function for the latent functions have \rbfKernel\ kernels
\cite{Alvarez:sparse2009}, or when the smoothing kernels have an
\rbfKernelLong\ form and the latent functions are Gaussian white noise
processes \cite{Pacioreck:nonstationaryCov:2004,
  Boyle:dependent04}. An alternative would be to consider discrete
process convolutions \cite{Higdon:convolutions02} instead of the
continuous process convolution of equations \eqref{eq:covf} and
\eqref{eq:covfq}, avoiding in this way the need to solve double
integrals.

We now briefly summarize different methods for reducing computational
complexity in multi-output Gaussian processes.

As we mentioned before, Rougier \cite{Rougier:multi08} assumes that
the multiple output problem can be seen as a single output problem
considering the output index as another variable of the input
space. The predicted output, $f(\boldx_*)$ is expressed as a weighted
sum of $Q$ deterministic regressors that explain the mean of the
output process plus a Gaussian error term that explains the variance
in the output. Both, the set of regressors and the covariance for the
error are assumed to be separable in the input space. The covariance
takes the form $k(\boldx,\boldx')k_T(d,d')$, as in the introduction of
section \ref{SepKer}. For isotopic models (\cite{Rougier:multi08}
refers to this condition as regular outputs, meaning outputs that are
evaluated at the same set of inputs $\boldX$), the mean and covariance
for the output, can be obtained through Kronecker products for the
regressors and the covariances involved in the error term.  For
inference the inversion of the necessary terms is accomplished using
properties of the Kronecker product. For example, if $\boldK(\boldX,
\boldX') = \boldB \otimes k(\boldX, \boldX')$, then
$\boldK^{-1}(\boldX, \boldX') = \boldB^{-1} \otimes k^{-1}(\boldX,
\boldX')$. Computational complexity is reduced to $O(D^3)+O(N^3)$,
similar to the eigendecomposition method in section
\ref{parEstimation:reg}.

Ver Hoef and Barry \cite{verHoelf:convolutionFFT04} present a
simulation example with $D=2$. Prediction over one of the variables is
performed using \emph{cokriging}. In cokriging scenarios, usually one
has access to a few measurements of a primary variable, but plenty of
observations for a secondary variable. In geostatistics, for example,
predicting the concentration of heavy pollutant metals (say Cadmium or
Lead), which are expensive to measure, can be done using inexpensive
and oversampled variables as a proxy (say pH levels)
\cite{Goovaerts:book97}. Following a suggestion by
\cite{Stein:Interpolation:1999} (page 172), the authors partition the
secondary observations into subgroups of observations and assume the
likelihood function is the sum of the partial likelihood functions of
several systems that include the primary observations and each of the
subgroups of the secondary observations. In other words, the joint
probability distribution $p(f_1(\boldX_1),f_2(\boldX_2))$ is
factorised as
$p(f_1(\boldX_1),f_2(\boldX_2))=\prod_{j=1}^{J}p(f_1(\boldX_1),f^{(j)}_2(\boldX^{(j)}_2))$,
where $f^{(j)}_2(\boldX^{(j)}_2)$ indicates the observations in the
subgroup $j$ out of $J$ subgroups of observations, for the secondary
variable. Inversion of the particular covariance matrix derived from
these assumptions grows as $O(JN^3)$, where $N$ is the number of input
points per secondary variable.

Also, the authors use a fast Fourier transform for computing the autocovariance matrices
$(\boldK(\boldX_d,\boldX_{d}))_{d,d}$ and cross-covariance matrices $(\boldK(\boldX_d,\boldX_{d'}))_{d,d'}$.

Boyle \cite{Boyle:phdThesis:2007} proposed an extension of the
\emph{reduced rank approximation} method presented by
\cite{Quinonero:reducedRank:2005}, to be applied to the dependent
Gaussian process construction.  The author outlined the generalization
of the methodology for $D=2$.  The outputs $f_1(\boldX_1)$ and
$f_2(\boldX_2)$ are defined as
\begin{align*}
\begin{bmatrix}
f_1(\boldX_1)\\
f_2(\boldX_2)
\end{bmatrix}&=
\begin{bmatrix}
(\boldK(\boldX_1,\boldX_1))_{1,1}& (\boldK(\boldX_1,\boldX_2))_{1,2}\\
(\boldK(\boldX_2,\boldX_1))_{2,1}& (\boldK(\boldX_2,\boldX_2))_{2,2}
\end{bmatrix}
\begin{bmatrix}
\widetilde{\boldw}_1\\
\widetilde{\boldw}_2
\end{bmatrix},
\end{align*}
where $\widetilde{\boldw}_d$ are vectors of weights associated to each
output including additional weights corresponding to the test inputs,
one for each output. Based on this likelihood, a predictive
distribution for the joint prediction of $f_1(\boldX)$ and
$f_2(\boldX)$ can be obtained, with the characteristic that the
variance for the approximation, approaches to the variance of the full
predictive distribution of the Gaussian process, even for test points
away from the training data. The elements in the matrices
$(\boldK(\boldX_d,\boldX_{d'}))_{d,d'}$ are computed using the
covariances and cross-covariances developed in sections \ref{SepKer}
and \ref{NonSepKer}. Computational complexity reduces to $O(DNM^2)$,
where $N$ is the number of sample points per output and $M$ is an user
specified value that accounts for the rank of the approximation.

In \cite{Alvarez:sparse2009}, the authors show how through making specific conditional independence assumptions, inspired
by the model structure in the process convolution formulation (for which the LMC is a special case), it is possible
to arrive at a series of efficient approximations that represent the covariance matrix $\Kmv$ using a reduced rank
approximation $\mathbf{Q}$ plus a matrix $\mathbf{D}$, where $\mathbf{D}$ has a specific structure that depends on the
particular independence
assumption made to obtain the approximation. Approximations can reduce the computational complexity to $O(NDM^2)$ with
$M$ representing a user specified value that determines the rank of $\mathbf{Q}$. Approximations
obtained in this way, have similarities with the conditional approximations summarized for a single output in
\cite{Quinonero:unifying05}.

Finally, the informative vector machine (IVM) \cite{Lawrence:ivm02}
has also been extended to Gaussian processes using kernel matrices
derived from particular versions of the linear model of
coregionalization, including \cite{Teh:semiparametric05} and
\cite{Lawrence:learningToLearn:2004}. In the IVM, only a smaller
subset of size $M$ of the data points is chosen for constructing the
GP predictor. The data points selected are the ones that maximize a
differential entropy score \cite{Lawrence:learningToLearn:2004} or an
information gain criteria \cite{Teh:semiparametric05}. Computational
complexity for this approximation is again $O(NDM^2)$.

For the computational complexities shown above, we assumed $R_q=1$ and $Q=1$.


\section{Applications of Multivariate Kernels}\label{applications}

In this chapter we further describe in more detail some of the
applications of kernel approaches to multi-output learning from
the statistics and machine learning communities.

One of the main application areas of multivariate Gaussian process has
been in computer emulation. In \cite{Fricker:FRFstructure2011}, the
LMC is used as the covariance function for a Gaussian process emulator
of a finite-element method that solves for frequency response
functions obtained from a structure. The outputs correspond to pairs
of masses and stiffnesses for several structural modes of vibration
for an aircraft model. The input space is made of variables related to
physical properties, such as Tail tip mass or Wingtip mass, among
others.

Multivariate computer emulators are also frequently used for modelling
time series. We mentioned this type of application in section
\ref{subsection:LMC:statistics:ML}. Mostly, the number of time points
in the time series are matched to the number of outputs (we expressed
this as $D=T$ before), and different time series correspond to
different input values for the emulation. The particular input values
employed are obtained from different ranges that the input variables
can take (given by an expert), and are chosen according to some
space-filling criteria (Latin hypercube design, for example)
\cite{Santner:ComputerExperimentsBook2003}. In \cite{Conti:multi09},
the time series correspond to the evolution of the net biome
productivity (NBP) index, which in turn is the output of the Sheffield dynamic
global vegetation model. In
\cite{McFarland:multiEmulator:Single:2008}, the time series is the
temperature of a particular location of a container with decomposing
foam. The simulation model is a finite element model and simulates the
transfer of heat through decomposing foam.

In machine learning the range of applications for multivariate kernels
is increasing. In \cite{Rogers:towards08}, the ICM is used to model
the dependencies of multivariate time series in a sensor
network. Sensors located in the south coast of England measure
different environmental variables such as temperature, wind speed,
tide height, among others. Sensors located close to each other make
similar readings. If there are faulty sensors, their missing readings
could be interpolated using the healthy ones.

In \cite{KianMing:MTGPRobotics:2009}, the authors use the ICM for
obtaining the inverse dynamics of a robotic manipulator.  The inverse
dynamics problem consists in computing the torques at different joints
of the robotic arm, as function of the angle, angle velocity and angle
acceleration for the different joints. Computed torques are necessary
to drive the robotic arm along a particular trajectory. Furthermore,
the authors consider several \emph{contexts}, this is, different
dynamics due to different loadings at the end effector. Joints are
modelled independently using an ICM for each of them, being the
outputs the different contexts and being the inputs, the angles, the
angle velocities and the angle accelerations. Besides interpolation,
the model is also used for extrapolation of novel contexts.

The authors of \cite{Bonilla:preferenceICM11} use the ICM for
preference elicitation, where a user is prompted to solve simple
queries in order to receive a recommendation. The ICM is used as a
covariance function for a GP that captures dependencies between users
(through the matrix $\boldB$), and dependencies between items (through
the covariance $k(\boldx,\boldx')$).

In \cite{Lawrence:gpsim2007a} and \cite{Gao:latent08}, the authors use
a process convolution to model the interaction between several genes
and a transcription factor protein, in a gene regulatory network. Each
output corresponds to a gene, and each latent function corresponds to
a transcription factor protein. It is assummed that transcription
factors regulate the rate at which particular genes produce primary
RNA. The output functions and the latent functions are indexed by
time. The smoothing kernel functions $G_{d,q}^{i}(\cdot)$ correspond
to the impulse response obtained from an ordinary differential
equation of first order. Given gene expression data, the problem is to
infer the time evolution of the transcription factor.

In \cite{Alvarez:lfm09}, the authors use a process convolution to
model the dependencies between different body parts of an actor that
performs modern dancing movements. This type of data is usually known
as mocap (for motion capture) data.  The outputs correspond to time
courses of angles referenced to a root node, for each body part
modelled. The smoothing kernel used corresponds to a Green's function
arising from a second order ordinary differential equation.

In \cite{RMurray:transformationGPs:2005}, the authors use a
discretized process convolution for solving an inverse problem in
reconstruction of images, and for fusing the information from multiple
sensors.

In \cite{Calder:randomWalks08}, two particulate matter (PM) levels
measured in the air (10 $\mu$m in diameter and 2.5 $\mu$m in
diameter), at different spatial locations, are modeled as the added
influence of coarse and fine particles. In turn, these coarse and fine
particles are modeled as random walks and then transformed by discrete
convolutions to represent the levels of $\text{PM}$ at $10$ $\mu$m and
$2.5$ $\mu$m.  The objective is to extract information about
$\text{PM}$ at $2.5$ $\mu$m from the abundant readings of $\text{PM}$
at $10$ $\mu$m.


\section{Discussion}\label{Conc}

We have presented a survey of multiple output kernel functions to be
used in kernel methods including regularization theory and Gaussian
processes. From the regularization theory point of view, the multiple
output problem can be seen as a regularization method that minimizes
directly a loss function while constraining the parameters of the
learned vector-valued function.  In a Gaussian process framework, from
a machine learning context, the multiple output problem is equivalent
to formulate a generative model for each output that expresses
correlations as functions of the output function index and the input
space, using a set of common latent functions.

We presented two general families of kernels for vector-valued
functions including the separable family (including the SoS kernels) and different instantiations
of what we would call the nonseparable family. The separable family
represent the kernel function as the product of a kernel for the
outputs, independently of the value that the input can have, and a
kernel function for the input space, independently of the output
index. The most general model is the linear model of
coregionalization, with many other kernel functions that appear in the
machine learning literature as particular cases. Within the family of
nonseparable kernels, the process convolution construction has proved
useful for several theoretical and applied problems and as we have
seen before, it can be considered as a generalization of the linear
model of coregionalization.

Model selection establishes a path for future research in
multiple-output kernels related problems. From a Bayesian perspective,
in the setup of LMC and process convolutions, model selection includes
principled mechanisms to find the number of latent functions and/or
the rank of the coregionalization matrices. More general model
selection problems involve the ability to test if given some data, the
outputs are really correlated or influence each other, compared to the
simpler assumption of independence. Other model selection problem
includes the influence of the input space configuration (isotopic
against heterotopic) towards the sharing of strengths between
outputs. Although such problems have been studied to some extent in
the geostatistics literature, there remain open issues.

\appendix

\section*{Acknowledgements}

  MA and NL are very grateful for support from a Google Research Award
  ``Mechanistically Inspired Convolution Processes for Learning'' and
  the EPSRC Grant No EP/F005687/1 ``Gaussian Processes for Systems
  Identification with Applications in Systems Biology''. MA also
  acknowledges the support from the Overseas Research Student Award
  Scheme (ORSAS), from the School of Computer Science of the
  University of Manchester and from the Universidad Tecnológica de
  Pereira, Colombia. LR would like to thank Luca Baldassarre and
  Tomaso Poggio for many useful discussions. LR is assistant professor at DISI,
  University of Genova and currently on leave of absence.
  We also thank to two anonymous reviewers for their helpful comments.

  This work was supported in part by the IST Programme of the European
  Community, under the PASCAL2 Network of Excellence,
  IST-2007-216886. Thanks to PASCAL 2 support the authors of this
  paper organized two workshops: \emph{Statistics and Machine Learning
    Interface Meeting} (see
  \url{http://intranet.cs.man.ac.uk/mlo/slim09/}), across 23-24 of
  July, 2009 at Manchester, UK and \emph{Kernels for Multiple Outputs
    and Multi-task Learning: Frequentist and Bayesian Points of View}
  (see \url{http://intranet.cs.man.ac.uk/mlo/mock09/}) held on
  December 12 at Whistler, Canada as part as one of the Workshops of
  NIPS 2009. This publication only reflects the authors' views, but
  they benefited greatly by interactions with other researchers at
  these workshops who included Andreas Argyriou, David Higdon, Tom
  Fricker, Sayan Mukherjee, Tony O'Hagan, Ian Vernon, Hans
  Wackernagel, Richard Wilkinson, and Chris Williams.

\newpage

\section*{Notation}
 \textbf{Generalities} \\
  \\
 $p\quad$ dimensionality of the input space \\
 $D\quad$  number of outputs\\
 $N, N_d\quad$ number of data points for output $d$ \\
 $Q\quad$  number of latent functions (for generative models) \\
 $\inputSpace\quad$  input space \\
 $\boldX_d\quad$   input training data for output $d$, $\boldX_d=\{\boldx_{d,n}\}_{n=1}^{N_d}$\\
 $\boldX\quad$  input training data for all outputs, $\boldX=\{\boldX_{d}\}_{d=1}^{D}$\\
  \\
 \textbf{Functions}  \\
   \\
 $k(\boldx, \boldx')\quad$ general scalar kernel\\
  $\boldK(\boldx, \boldx')\quad$  general kernel valued matrix with entries $\left(\boldK(\boldx, \boldx')\right)_{d,d'}$ with $d,d=1, \dots, D$\\
 $k_q(\boldx, \boldx')\quad$ scalar kernel for the $q-$th latent function\\
 $f_d(\boldx)\quad$  $d$-th output evaluated at $\boldx$\\
 $\boldf(\boldx)\quad$,  vector-valued function, $\boldf(\boldx)= [f_1(\boldx), \ldots,f_D(\boldx) ]^{\top}$\\
 $\delta_{k,k'}\quad$  Kronecker delta for discrete arguments\\
 $\delta(x)\quad$     Dirac delta for continuous arguments\\
  \\
 \textbf{Vectors and matrices}  \\
   \\
 $k_q(\boldX, \boldX)\quad$  kernel matrix with entries $k_{q}(\boldx,\boldx')$ evaluated at $\boldX$\\
 $\boldf_d(\boldX_d)\quad$  $\boldf_d(\boldx)$ evaluated at $\boldX_d$, $\boldf_d=[f_d(\boldx_{d,1}), \ldots, f_d(\boldx_{d,N_d})]^{\top}$\\
 $\boldf(\boldX)\quad$  vectors $\{\boldf_d\}_{d=1}^D$, stacked in a column vector\\
 $\left(\boldK(\boldX_d,\boldX_{d'})\right)_{d,d'}\quad$  kernel matrix with entries $(\boldK(\boldx_{d,n},\boldx_{d',m}))_{d,d'}$
with $\boldx_{d,n}\in \boldX_d$ and $\boldx_{d',m}\in \boldX_{d'}$\\
$\boldK(\boldX, \boldX)\quad$  kernel matrix with blocks $\left(\boldK(\boldX_d,\boldX_{d'})\right)_{d,d'}$ with
$d,d'=1, \dots, D\quad$\\
 $\boldI_{N}\quad$  identity matrix of size $N$\\

\end{document}